\documentclass{article}

\PassOptionsToPackage{numbers, compress}{natbib}
\usepackage[preprint]{neurips_2022}

\usepackage[utf8]{inputenc} % allow utf-8 input
\usepackage[T1]{fontenc}    % use 8-bit T1 fonts
\usepackage{hyperref}       % hyperlinks
\usepackage{url}            % simple URL typesetting
\usepackage{booktabs}       % professional-quality tables
\usepackage{amsfonts}       % blackboard math symbols
\usepackage{nicefrac}       % compact symbols for 1/2, etc.
\usepackage{microtype}      % microtypography
\usepackage[table]{xcolor}         % colors
\usepackage{graphicx}
\usepackage{multirow}
\usepackage{subcaption}
\usepackage{amsmath}
\usepackage[capitalise]{cleveref}
\usepackage{xspace}
\usepackage{caption}
\usepackage{subcaption}
\usepackage{pifont}
\usepackage{arydshln}
\usepackage{enumitem}
\usepackage[outline]{contour}
\usepackage{comment}

\usepackage{tikz}
\usetikzlibrary{decorations.pathreplacing, calligraphy}
\usetikzlibrary{shapes.callouts}
\usetikzlibrary{chains, fit, quotes, automata}
\usetikzlibrary{arrows}
\usetikzlibrary{matrix,positioning}
\usetikzlibrary{shapes.geometric, shapes.misc}
\usetikzlibrary{calc}

\definecolor{yesColor}{RGB}{37,37,37}
\definecolor{noColor}{RGB}{189,189,189}
\newcommand{\cmark}{\textcolor{yesColor}{\ding{51}}}%
\newcommand{\xmark}{\textcolor{noColor}{\ding{55}}}%

\definecolor{mv1}{HTML}{d7191c}
\definecolor{mv2}{HTML}{4dac26}

\newcommand{\arch}{\texttt{MobileViTv2}\xspace}

\title{Separable Self-attention for Mobile Vision Transformers}

\author{%
  Sachin Mehta \\
  Apple \\
  \And 
  Mohammad Rastegari \\
  Apple\\
}

\begin{document}

\maketitle

\begin{abstract}
  Mobile vision transformers (MobileViT) can achieve state-of-the-art performance across several mobile vision tasks, including classification and detection. Though these models have fewer parameters, they have high latency as compared to convolutional neural network-based models. The main efficiency bottleneck in MobileViT is the multi-headed self-attention (MHA) in transformers, which requires $O(k^2)$ time complexity with respect to the number of tokens (or patches) $k$. Moreover, MHA requires costly operations (e.g., batch-wise matrix multiplication) for computing self-attention, impacting latency on resource-constrained devices. This paper introduces a \emph{separable self-attention} method with linear complexity, i.e. $O(k)$. A simple yet effective characteristic of the proposed method is that it uses element-wise operations for computing self-attention, making it a good choice for resource-constrained devices. The improved model, \arch, is state-of-the-art on several mobile vision tasks, including ImageNet object classification and MS-COCO object detection. With about three million parameters, \arch~achieves a top-1 accuracy of 75.6\% on the ImageNet dataset, outperforming MobileViT by about 1\% while running $3.2\times$ faster on a mobile device. Our source code is available at: \url{https://github.com/apple/ml-cvnets}
\end{abstract}

\section{Introduction}
\begin{figure}[t!]
    \centering
    \includegraphics[width=0.8\columnwidth]{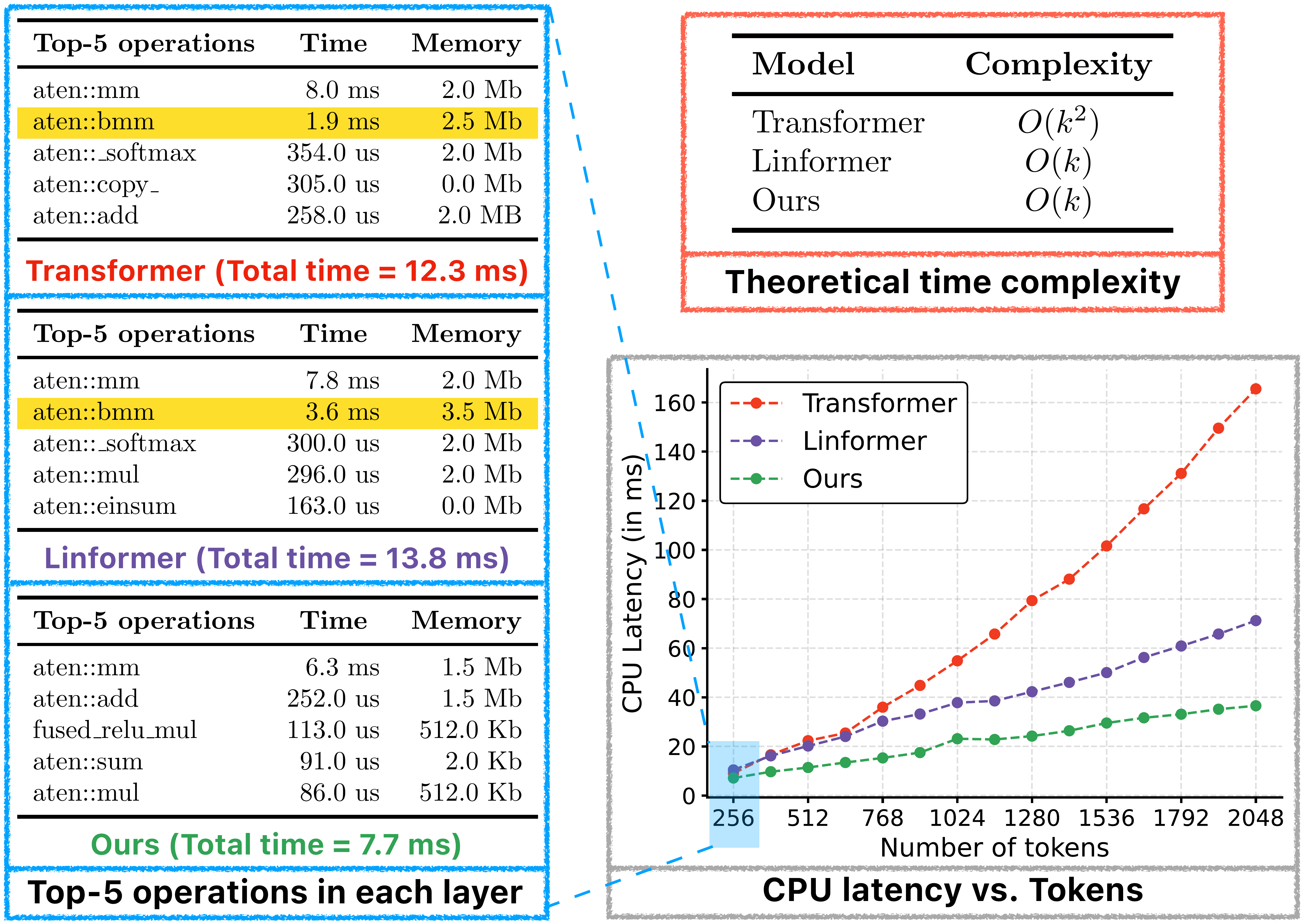}
    \caption{\textbf{Comparison between different attention units.} Transformer and Linformer use costly operations (batch-wise matrix multiplication) for computing self-attention. Such operations are a bottleneck for efficient inference on resource-constrained devices. The proposed method does not use such operations, thus accelerating inference on resource-constrained devices. \textbf{Left} compares top-5 operations (sorted by CPU time) in a single layer of different attention units for $k=256$ tokens. \textbf{Top Right} compares complexity of different attention units. \textbf{Bottom Right} compares the latency of different attention units as a function of the number of tokens $k$. These results are computed on a single CPU core machine with a 2.4 GHz 8-Core Intel Core i9 processor, $d=512$ (token dimensionality), $h=8$ (number of heads; for Transformer and Linformer), and $p=256$ (projected tokens in Linformer) using a publicly available profiler in PyTorch \cite{paszke2019pytorch}.}
    \label{fig:compare_different_attn_layers}
\end{figure}

Vision transformers (ViTs) \cite{dosovitskiy2021an} have become ubiquitous for a wide variety of visual recognition tasks \cite{touvron2021training,liu2021swin}, including mobile vision tasks \cite{mehta2022mobilevit}. At the heart of the ViT-based models, including mobile vision transformers, is the transformer block \cite{vaswani2017attention}. The main efficiency bottleneck in ViT-based models, especially for inference on resource-constrained devices, is the multi-headed self-attention (MHA). MHA allows the tokens (or patches) to interact with each other, and is a key for learning global representations. However, the complexity of self-attention in transformer block is $O(k^2)$, i.e., it is quadratic with respect to the number of tokens (or patches) $k$. Besides this, computationally expensive operations (e.g., batch-wise matrix multiplication; see  \cref{fig:compare_different_attn_layers}) are required to compute attention matrix in MHA. This, in particular, is concerning for deploying ViT-based models on resource-constrained devices, as these devices have reduced computational capabilities, restrictive memory constraints, and a limited power budget. Therefore, this paper seeks to answer this question: \emph{can self-attention in transformer block be optimized for resource-constrained devices?}

\begin{figure}[t!]
    \centering
    \begin{subfigure}[b]{0.32\columnwidth}
        \centering
        \includegraphics[width=\columnwidth]{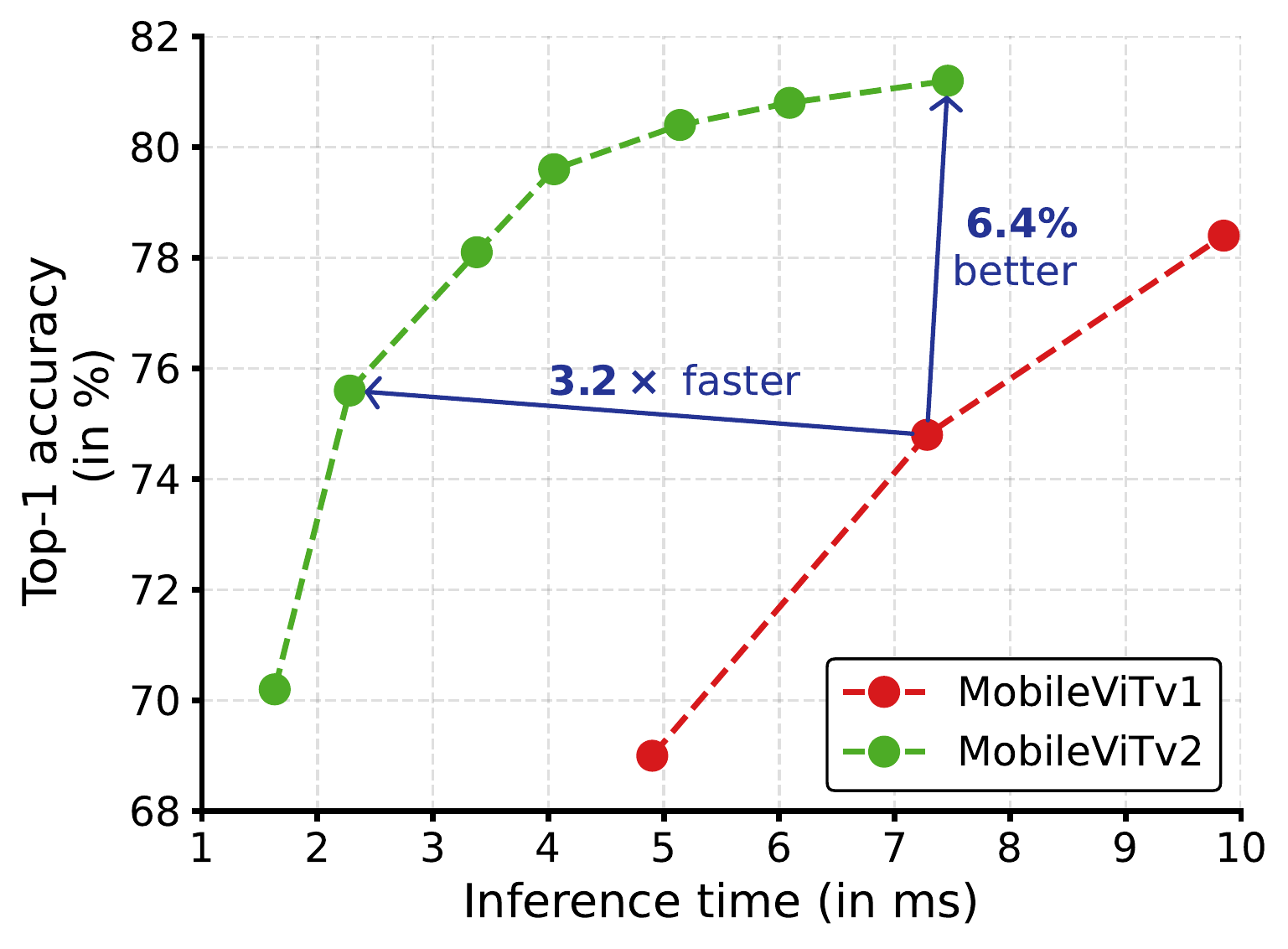}
        \caption{ImageNet-1k classification}
        \label{fig:compre_mv1_mv2_in1k}
    \end{subfigure}
    \hfill
    \begin{subfigure}[b]{0.32\columnwidth}
        \centering
        \includegraphics[width=\columnwidth]{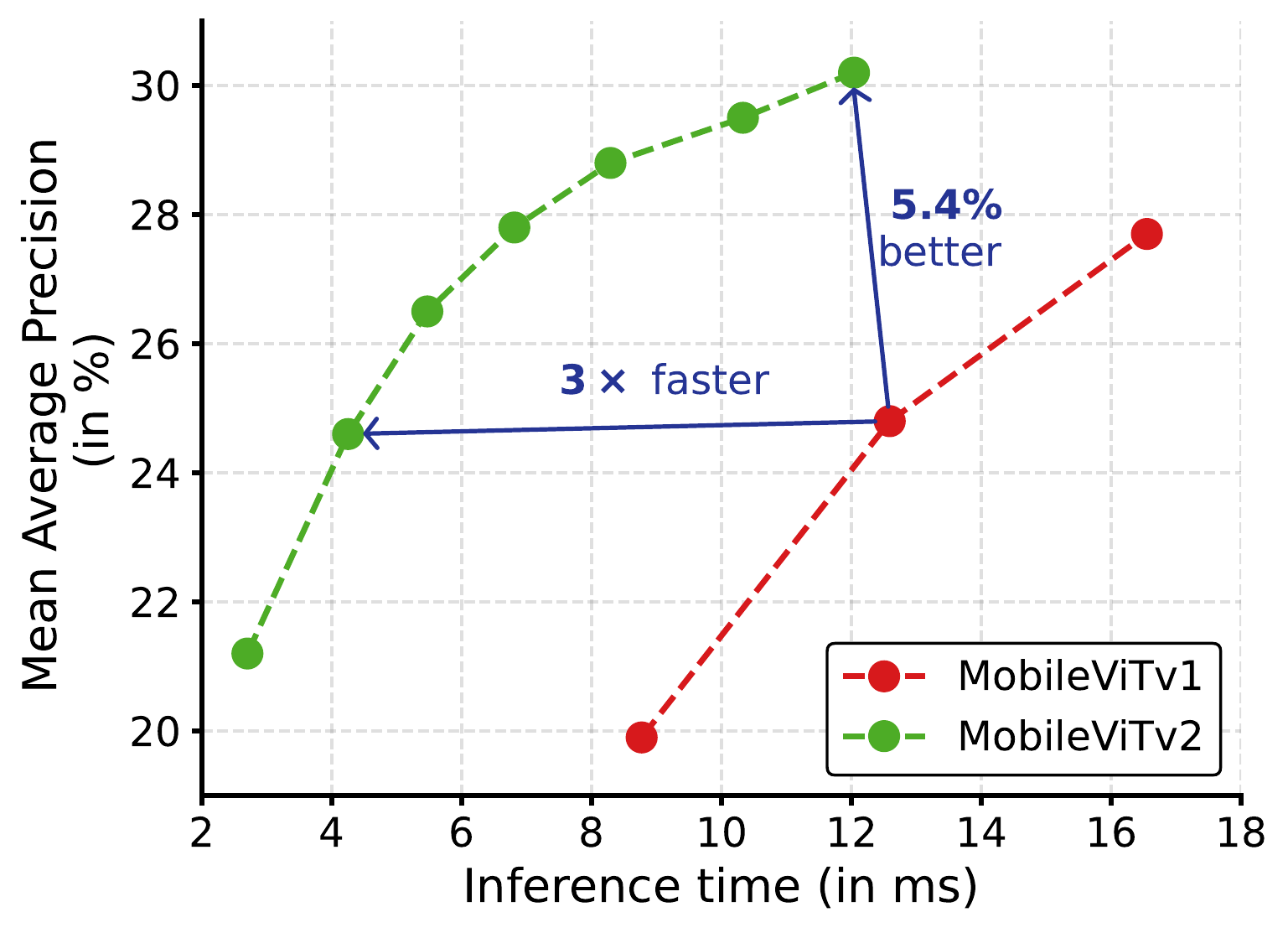}
        \caption{MS-COCO object detection}
        \label{fig:compre_mv1_mv2_ssd}
    \end{subfigure}
    \hfill
    \begin{subfigure}[b]{0.32\columnwidth}
        \centering
        \includegraphics[width=\columnwidth]{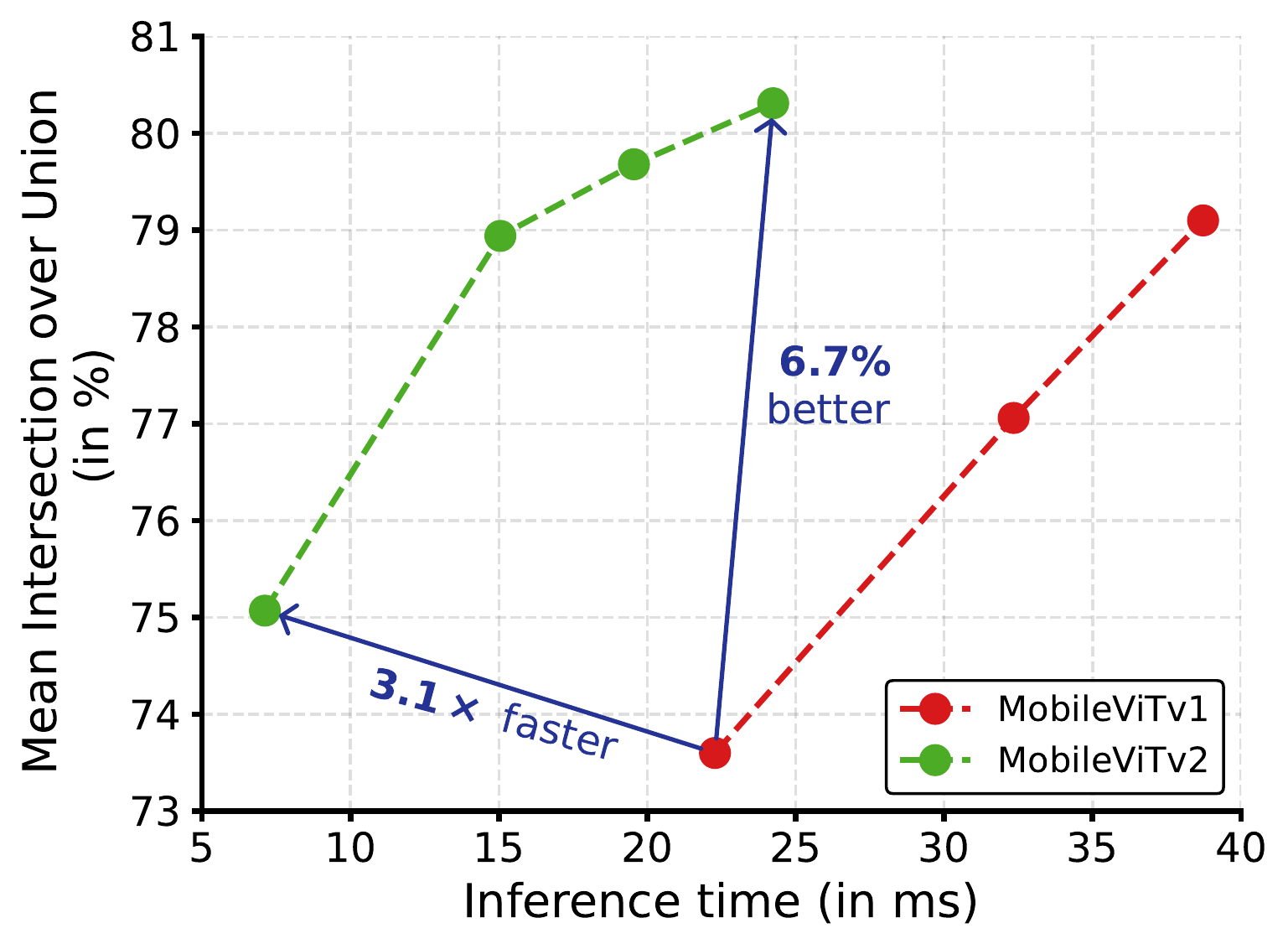}
        \caption{PASCAL VOC segmentation}
        \label{fig:compre_mv1_mv2_deeplab}
    \end{subfigure}
    \caption{\textbf{\textcolor{mv2}{\arch}~models are faster and better than \textcolor{mv1}{MobileViTv1} models \cite{mehta2022mobilevit} across different tasks.} \textcolor{mv2}{\arch}~models are constructed by replacing multi-headed self-attention in \textcolor{mv1}{MobileViTv1} with the proposed \emph{separable self-attention} (\cref{ssec:separable_self_attn}). Here, inference time is measured on an iPhone12 for an input resolution of $256 \times 256$, $512 \times 512$, and $320 \times 320$ for classification, segmentation, and detection respectively.}
    \label{fig:compare_mobilevitv1_mobilevitv2}
\end{figure}

Several methods \cite[e.g.,][]{child2019generating, kitaev2020reformer, beltagy2020longformer, wang2020linformer} have been proposed for optimizing the self-attention operation in transformers (not necessarily for ViTs). Among these, a widely studied approach in sequence modeling tasks is to introduce sparsity in self-attention layers, wherein each token  attends to a subset of tokens in an input sequence \cite{child2019generating, beltagy2020longformer}. Though these approaches reduces the time complexity from $O(k^2)$ to $O(k \sqrt{k})$ or $O(k \log{k})$, the cost is a performance drop. Another popular approach for approximating self-attention is via low-rank approximation. Linformer \cite{wang2020linformer} decomposes the self-attention operation into multiple smaller self-attention operations via linear projections, and reduces the complexity of self-attention from $O(k^2)$ to $O(k)$. However, Linformer still uses costly operations (e.g., batch-wise matrix multiplication; \cref{fig:compare_different_attn_layers}) for learning global representations in MHA, which may hinder the deployment of these models on resource-constrained devices. 

This paper introduces a novel method, \emph{separable self-attention}, with $O(k)$ complexity for addressing the bottlenecks in MHA in transformers. For efficient inference, the proposed self-attention method also replaces the computationally expensive operations (e.g., batch-wise matrix multiplication) in MHA with element-wise operations (e.g., summation and multiplication). Experimental results on standard vision datasets and tasks demonstrates the effectiveness of the proposed method (\cref{fig:compare_mobilevitv1_mobilevitv2}). 

\section{Related work}

\paragraph{Improving self-attention} Improving the efficiency of MHA in transformers is an active area of research. The first line of research introduces locality to address the computational bottleneck in MHA \cite[e.g.,][]{ child2019generating, beltagy2020longformer, parmar2018image, qiu2019blockwise}. Instead of attending to all $k$ tokens, these methods use predefined patterns to limit the receptive field of self-attention from all $k$ tokens to a subset of tokens, reducing the time complexity from $O(k^2)$ to $O(k \sqrt{k})$ or $O(k \log{k})$. However, such methods suffer from large performance degradation with moderate training/inference speed-up over the standard MHA in transformers. To improve the efficiency of MHA, the second line of research uses similarity measures to group tokens \cite{kitaev2020reformer, vyas2020fast, wang2021cluster}. For instance, Reformer \cite{kitaev2020reformer} uses locality-sensitive hashing to group the tokens and reduces the theoretical self-attention cost from $O(k^2)$ to $O(k \log{k})$. However, the efficiency gains over standard MHA are noticeable only for large sequences ($k > 2048$) \cite{kitaev2020reformer}. Because $k < 1024$ in ViTs, these approaches are not suitable for ViTs. The third line of research improves the efficiency of MHA via low-rank approximation \cite{wang2020linformer, choromanski2020rethinking}. The main idea is to approximate the self-attention matrix with a low-rank matrix, reducing the computational cost from $O(k^2)$ to $O(k)$. Even though these methods speed-up the self-attention operation significantly, they still use expensive operations for computing attention, which may hinder the deployment of these models on resource-constrained devices (\cref{fig:compare_different_attn_layers}). 

In summary, existing methods for improving MHA are limited in their reduction of inference time and memory consumption, especially for resource-constrained devices. This work introduces a separable self-attention method that is fast and memory-efficient (see \cref{fig:compare_different_attn_layers}), which is desirable for resource-constrained devices.

\paragraph{Improving transformer-based models} There has been  significant work on improving the efficiency of transformers \cite{liu2021swin, mehta2022mobilevit, mehta2021delight, wu2021cvt, heo2021rethinking}. The majority of these approaches reduce the number of tokens in the transformer block using different methods, including down-sampling \cite{ryoo2021tokenlearner,heo2021rethinking} and pyramidal structure \cite{liu2021swin, wang2021pyramid, mehta2022mobilevit}. Because the proposed separable self-attention module is a drop-in replacement to MHA, it can be easily integrated with any transformer-based model to further improve its efficiency.

\paragraph{Other methods} Transformer-based models performance can be improved using different methods, including mixed-precision training \cite{micikevicius2018mixed}, efficient optimizers \cite{dettmers2022bit, xiaohua2021scaling}, and knowledge distillation \cite{touvron2021training}. These methods are orthogonal to our work, and by default, we use mixed-precision during training.

\section{\arch}

\begin{figure}[t!]
    \centering
    \begin{subfigure}[b]{\columnwidth}
        \centering
        \includegraphics[width=0.8\columnwidth]{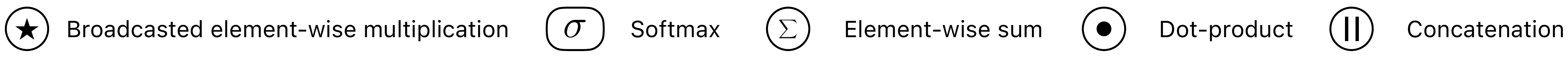}
    \end{subfigure}
    \vfill
    \begin{subfigure}[b]{0.3\columnwidth}
        \centering
        \includegraphics[height=140px]{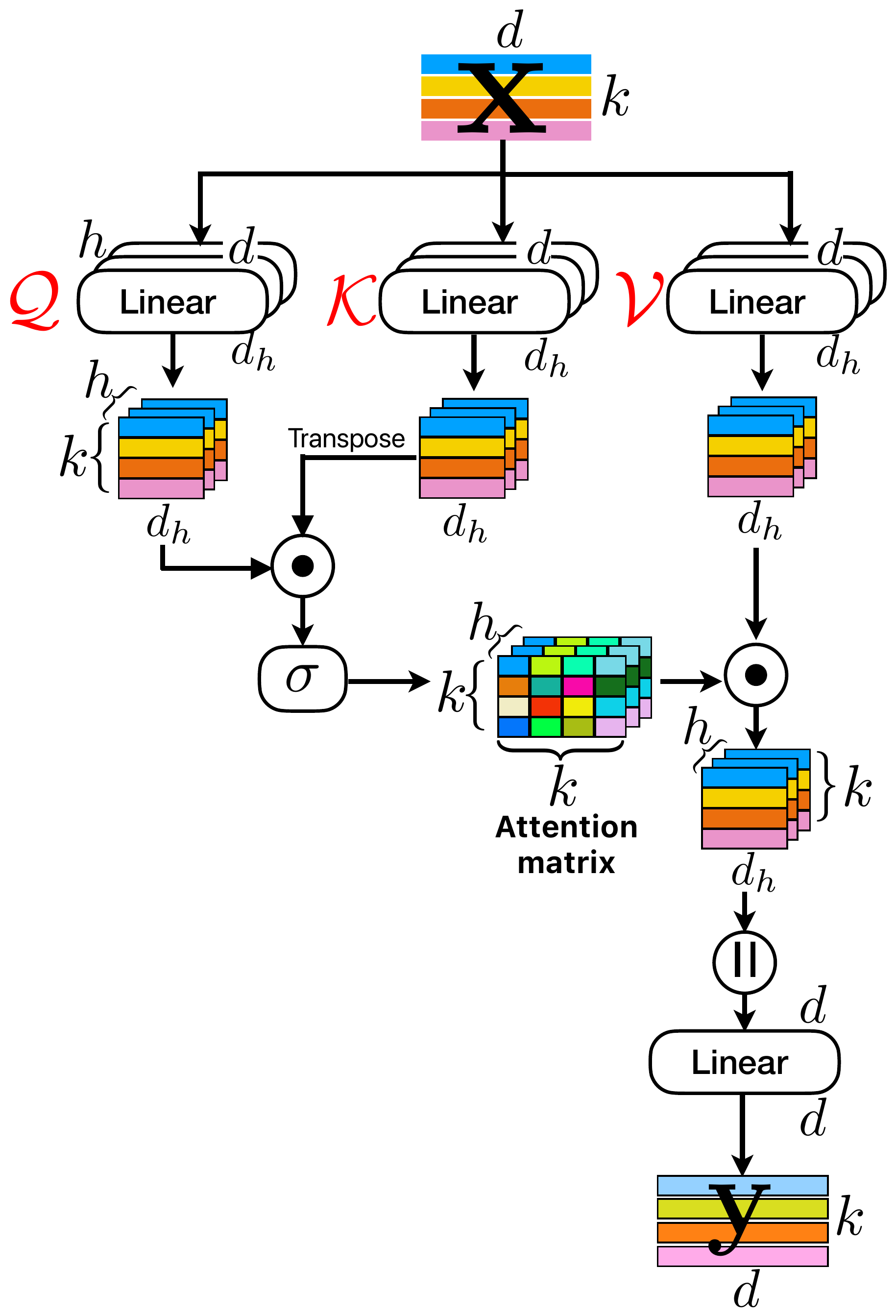}
        \caption{MHA in Transformers \cite{vaswani2017attention}}
        \label{fig:transformer_self_attn}
    \end{subfigure}
    \hfill
    \begin{subfigure}[b]{0.3\columnwidth}
        \centering
        \includegraphics[height=140px]{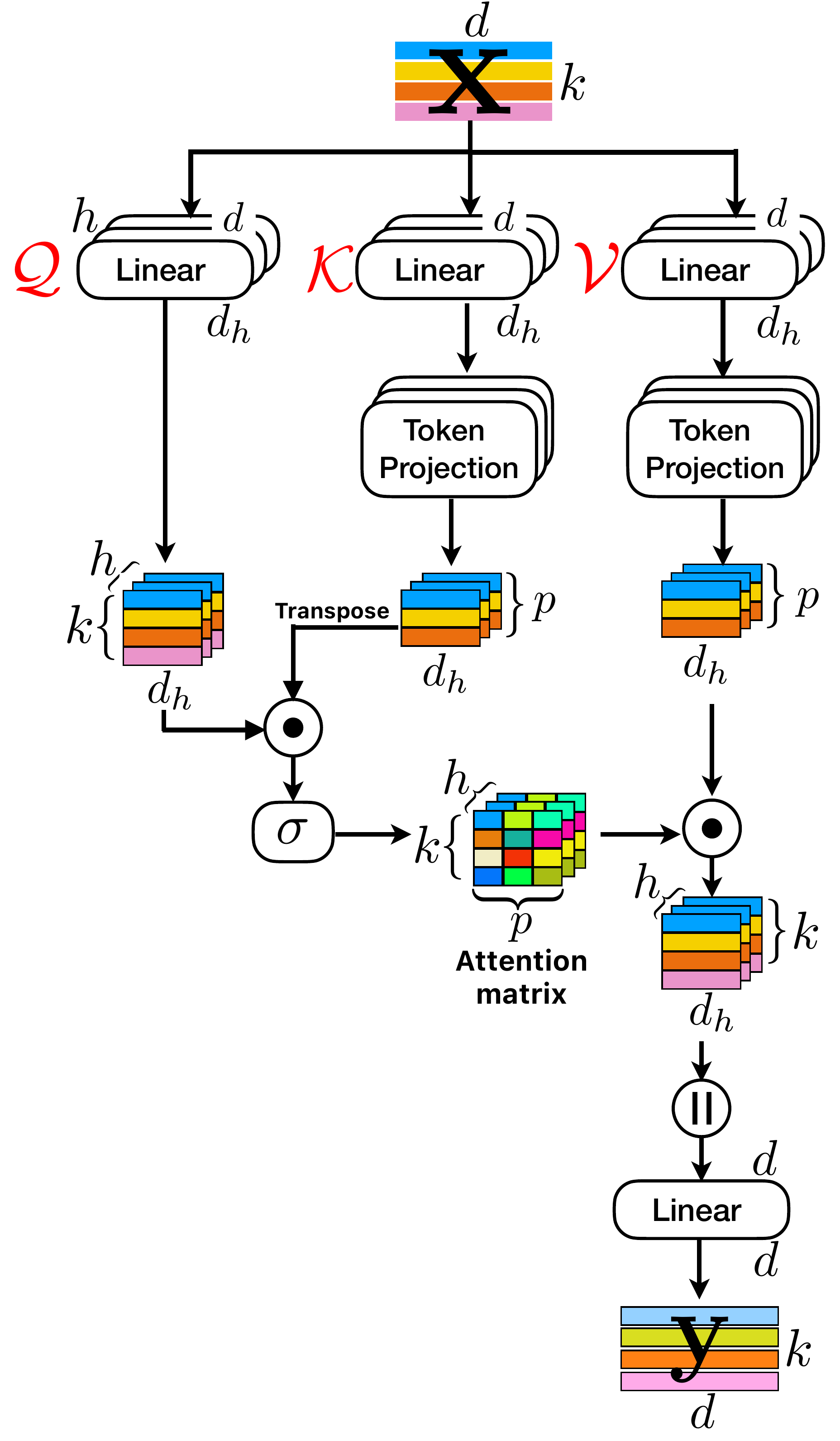}
        \caption{MHA in Linformer \cite{wang2020linformer}}
        \label{fig:linformer_self_attn}
    \end{subfigure}
    \hfill
    \begin{subfigure}[b]{0.3\columnwidth}
        \centering
        \includegraphics[height=140px]{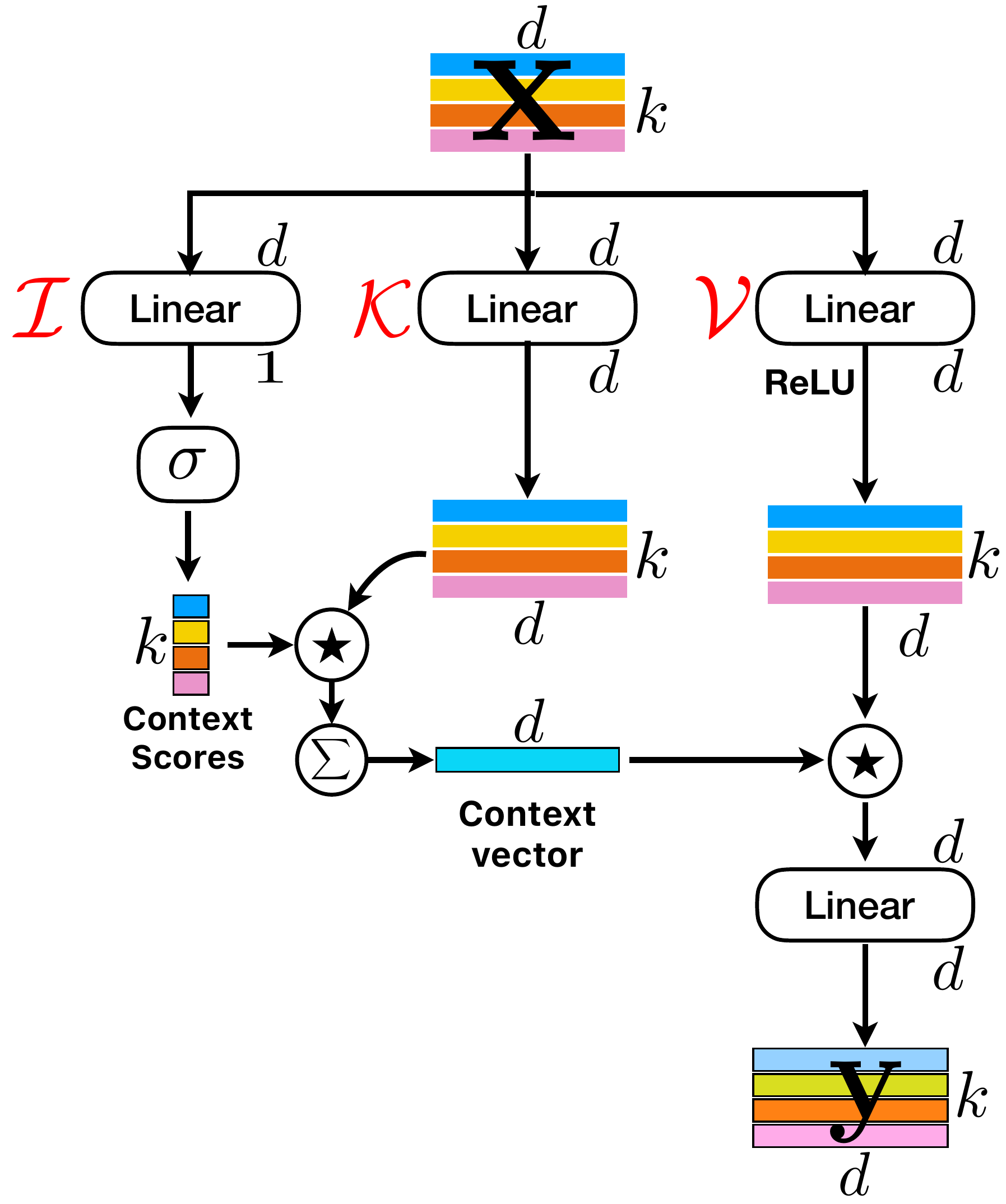}
        \caption{Separable self-attention (ours)}
        \label{fig:eff_self_attn}
    \end{subfigure}
    \caption{\textbf{Different self-attention units.} \textbf{(a)} is a standard multi-headed self-attention (MHA) in transformers. \textbf{(b)} extends MHA in (a) by introducing token projection layers, which project $k$ tokens to a pre-defined number of tokens $p$, thus reducing the complexity from $O(k^2)$ to $O(k)$. However, it still uses costly operations (e.g., batch-wise matrix multiplication) for computing self-attention, impacting latency on resource-constrained devices (\cref{fig:compare_different_attn_layers}). \textbf{(c)} is the proposed separable self-attention layer that is linear in complexity, i.e., $O(k)$, and uses element-wise operations for faster inference.}
    \label{fig:architecture}
\end{figure}

MobileViT \cite{mehta2022mobilevit} is a hybrid network that combines the strengths of CNNs and ViTs. MobileViT views transformers as convolutions, which allows it to leverage the merits of both convolutions (e.g., inductive biases) and transformers (e.g., long-range dependencies) to build a light-weight network for mobile devices. Though MobileViT networks have significantly fewer parameters and deliver better performance as compared to light-weight CNNs (e.g., MobileNets \cite{sandler2018mobilenetv2, howard2019searching}), they have high latency. The main efficiency bottleneck in MobileViT is the multi-headed self-attention (MHA; \cref{fig:transformer_self_attn}). 

MHA uses scaled dot-product attention to capture the contextual relationships between $k$ tokens (or patches). However, MHA is expensive as it has $O(k^2)$ time complexity. This quadratic cost is a bottleneck for transformers with a large number of tokens $k$ (\cref{fig:compare_different_attn_layers}). Moreover, MHA uses computationally- and memory-intensive operations (e.g., batch-wise matrix multiplication and softmax for computing attention matrix; \cref{fig:compare_different_attn_layers}); which could be a bottleneck on resource-constrained devices. To address the limitations of MHA for efficient inference on resource-constrained devices, this paper introduces \emph{separable self-attention} with linear complexity (\cref{fig:eff_self_attn}). 

\begin{figure}
    \centering
    \begin{subfigure}[b]{\columnwidth}
        \centering
        \includegraphics[width=0.7\columnwidth]{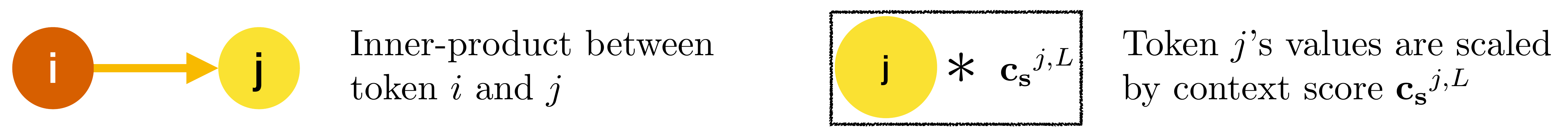}
        \vspace{4mm}
    \end{subfigure}
    \vfill
    \begin{subfigure}[b]{0.49\columnwidth}
        \centering
        \includegraphics[height=120px]{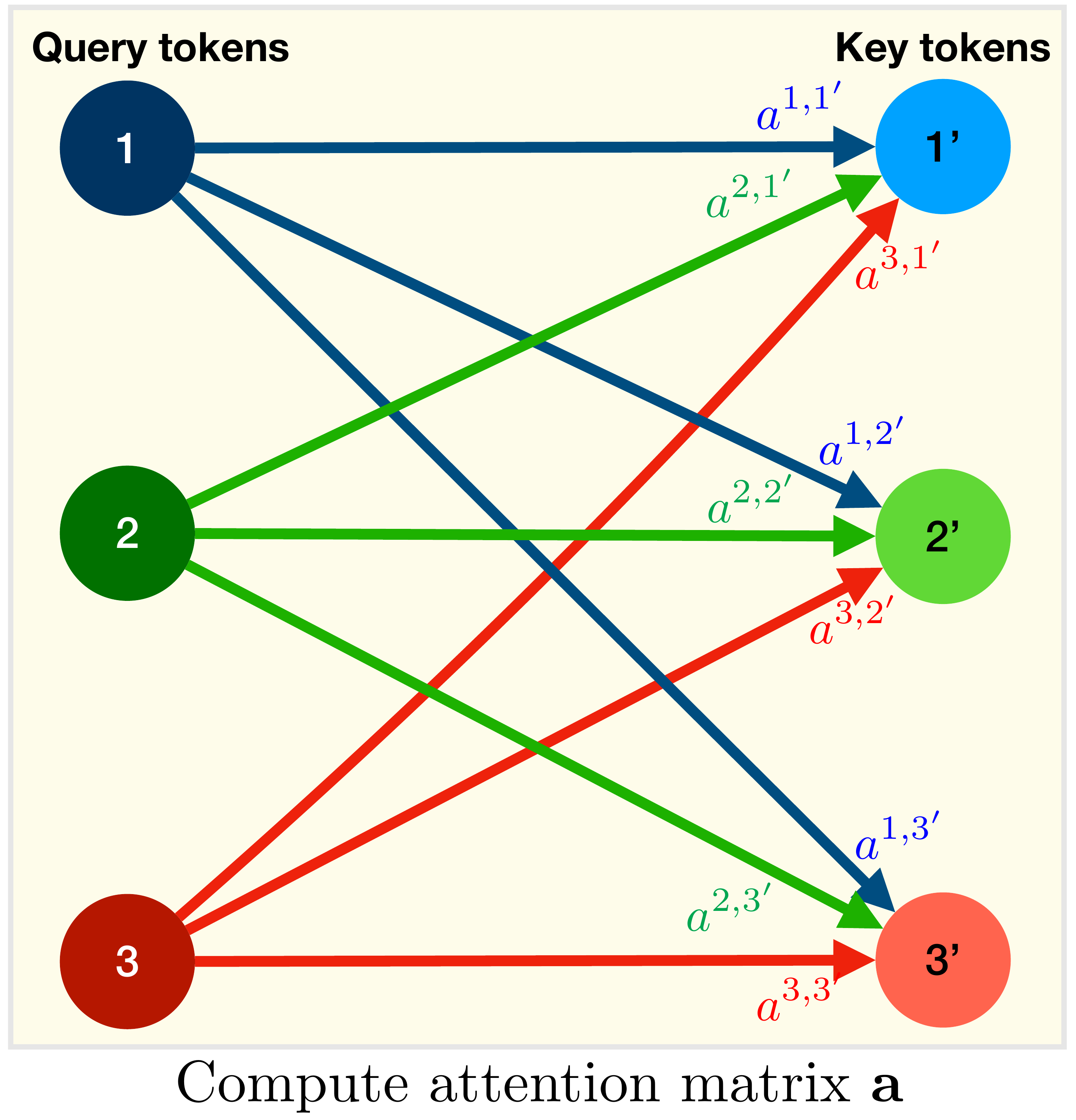}
        \caption{Self-attention in transformers}
        \label{fig:idea_self_attn}
    \end{subfigure}
    \hfill
    \begin{subfigure}[b]{0.49\columnwidth}
        \centering
        \includegraphics[height=120px]{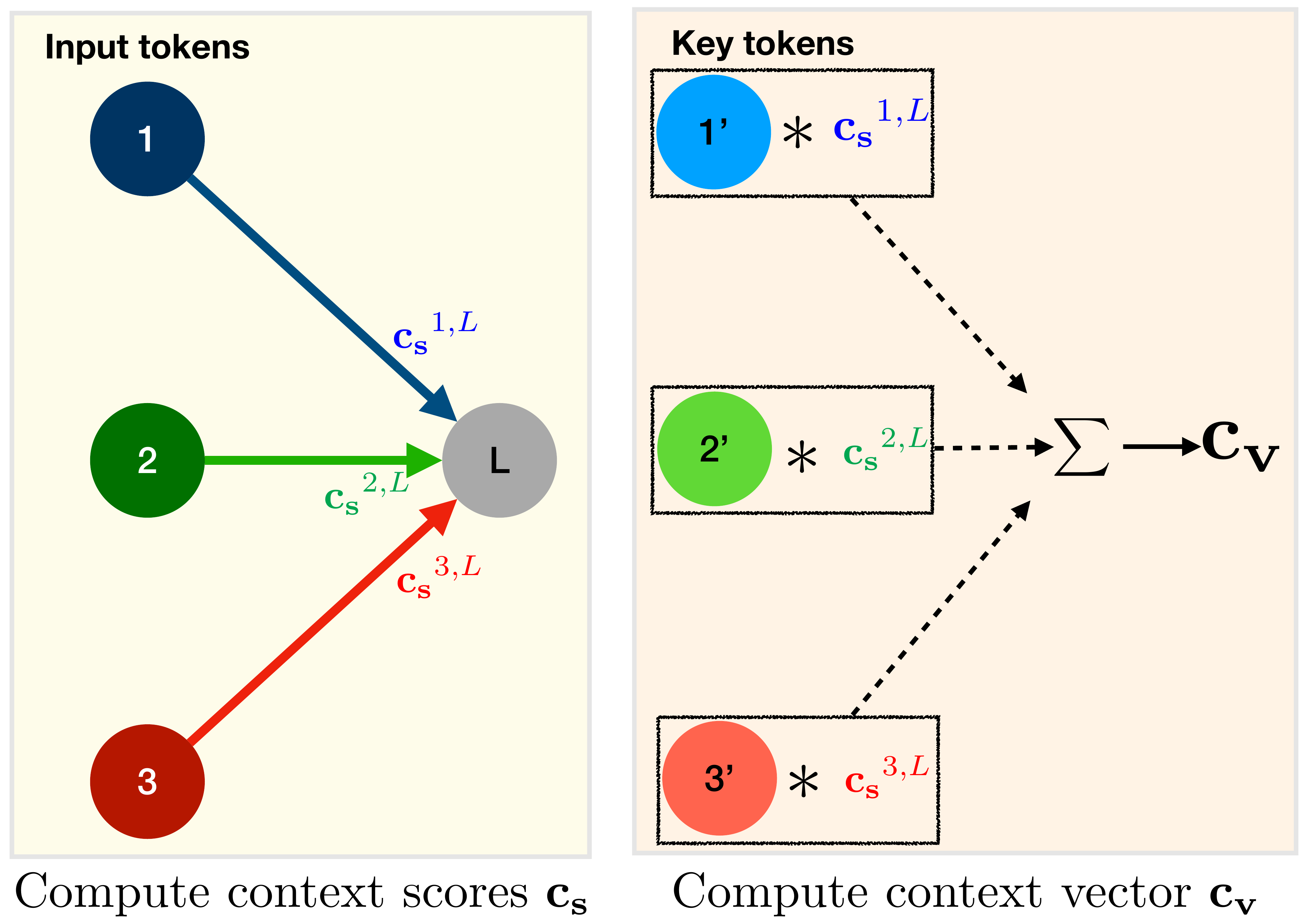}
        \caption{Proposed separable self-attention method}
        \label{fig:idea_sep_attn}
    \end{subfigure}
    \caption{\textbf{Example illustrating the interaction between tokens to learn global representations in different attention layers.} In (a), each query token computes the distance with all key tokens via dot-product. These distances are then normalized using softmax to produce an attention matrix $\mathbf{a}$, which encodes contextual relationships. In (b), the inner product between input tokens and latent token $L$ is computed. The resultant vector is normalized using softmax to produce context scores $\mathbf{c_s}$. These context scores are used to weight key tokens and produce a context vector $\mathbf{c_v}$, which encodes contextual information.}
    \label{fig:self_attn_compare_graph}
\end{figure}

The main idea of our \emph{separable self-attention} approach, shown in \cref{fig:idea_sep_attn}, is to compute context scores with respect to a latent token $L$. These scores are then used to re-weight the input tokens and produce a context vector, which encodes the global information. Because the self-attention is computed with respect to a latent token, the proposed method can reduce the complexity of self-attention in the transformer by a factor $k$. A simple yet effective characteristic of the proposed method is that it uses element-wise operations (e.g., summation and multiplication) for its implementation, making it a good choice for resource-constrained devices. We call the proposed attention method separable self-attention because it allows us to encode global information by replacing the quadratic MHA with two separate linear computations. The improved model, \arch, is obtained by replacing MHA with separable self-attention in MobileViT. 

In the rest of this section, we first briefly describe MHA (\cref{ssec:multi_head_attn}), and then elaborate on the details of separable self-attention (\cref{ssec:separable_self_attn}) and \arch architecture (\cref{ssec:mobilevitv2_arch}).

\subsection{Overview of multi-headed self-attention}
\label{ssec:multi_head_attn}
MHA (\cref{fig:transformer_self_attn}) allows transformer to encode inter-token relationships. Specifically, MHA takes an input $\mathbf{x} \in \mathbb{R}^{k \times d}$ comprising of $k$ $d$-dimensional token (or patch) embeddings. The input $\mathbf{x}$ is then fed to three branches, namely query $\mathcal{Q}$, key $\mathcal{K}$, and value $\mathcal{V}$. Each branch ($\mathcal{Q}$, $\mathcal{K}$, and $\mathcal{V}$) is comprised of $h$  linear layers (or heads), which enables the transformer to learn multiple views of the input. The dot-product between the output of linear layers in $\mathcal{Q}$ and $\mathcal{K}$ is then computed simultaneously for all $h$ heads, and is followed by a softmax operation $\sigma$ to produce an attention (or context-mapping) matrix $\mathbf{a} \in \mathbb{R}^{k \times k \times h}$. Another dot-product is then computed between $\mathbf{a}$ and the output of linear layers in $\mathcal{V}$ to produce weighted sum output $\mathbf{y_w} \in \mathbb{R}^{k \times d_h \times h}$, where $d_h =\frac{d}{h}$ is the head dimension . The outputs of $h$ heads are concatenated to produce a tensor with $k$ $d$-dimensional tokens, which is then fed to another linear layer with weights $\mathbf{W_O} \in \mathbb{R}^{d \times d}$ to produce the output of MHA $\mathbf{y} \in \mathbb{R}^{k \times d}$. Mathematically, this operation can be described as:
\begin{equation}
    \resizebox{0.85\hsize}{!}{
        $\mathbf{y} = \text{Concat} \left( \langle\ \underbrace{\sigma \left(\ \langle \mathbf{x} \mathbf{W_Q}^0, \mathbf{x} \mathbf{W_K}^0 \rangle\ \right)}_{\mathbf{a}^0 \in \mathbb{R}^{k \times k}}, \mathbf{x} \mathbf{W_V}^0\ \rangle, \cdots, \langle\ \underbrace{\sigma \left(\ \langle \mathbf{x} \mathbf{W_Q}^h, \mathbf{x} \mathbf{W_K}^h \rangle\ \right)}_{\mathbf{a}^h \in \mathbb{R}^{k \times k}}, \mathbf{x} \mathbf{W_V}^h\ \rangle \right) \mathbf{W_O}$
    }
    \label{eq:self_attn_transformer}
\end{equation}
where $\mathbf{W_Q}^i \in \mathbb{R}^{d \times d_h}$, $\mathbf{W_K}^i \in \mathbb{R}^{d \times d_h}$, and $\mathbf{W_V}^i \in \mathbb{R}^{d \times d_h}$ are the weights of the $i$-th linear layer (or head) in $\mathcal{Q}$, $\mathcal{K}$, and $\mathcal{V}$ branches respectively. The symbol $\langle \cdot, \cdot \rangle$ denotes the dot-product operation.

\subsection{Separable self-attention}
\label{ssec:separable_self_attn}
The structure of separable self-attention is inspired by MHA. Similar to MHA, the input $\mathbf{x}$ is processed using three branches, i.e., input $\mathcal{I}$, key $\mathcal{K}$, and value $\mathcal{V}$. The input branch $\mathcal{I}$ maps each $d$-dimensional token in $\mathbf{x}$ to a scalar using a linear layer with weights $\mathbf{W_I} \in \mathbb{R}^{d}$. The weights $\mathbf{W_I}$ serves as the latent node $L$ in \cref{fig:idea_sep_attn}. This linear projection is an inner-product operation and computes the distance between latent token $L$ and $\mathbf{x}$, resulting in a $k$-dimensional vector. A softmax operation is then applied to this $k$-dimensional vector to produce context scores $\mathbf{c_s} \in \mathbb{R}^{k}$. Unlike transformers that compute the attention (or context) score for each token with respect to all $k$ tokens, the proposed method only computes the context score with respect to a latent token $L$. This reduces the cost of computing attention (or context) scores from $O(k^2)$ to $O(k)$. 

The context scores $\mathbf{c_s}$ are used to compute a context vector $\mathbf{c_v}$. Specifically, the input $\mathbf{x}$ is linearly projected to a $d$-dimensional space using key branch $\mathcal{K}$ with weights $\mathbf{W_K} \in \mathbb{R}^{d \times d}$ to produce an output $\mathbf{x_K} \in \mathbb{R}^{k \times d}$. The context vector $\mathbf{c_v} \in \mathbb{R}^{d}$ is then computed as a weighted sum of $\mathbf{x_K}$ as: 
\begin{equation}
 \mathbf{c_v} = \sum_{i=1}^{k} \mathbf{c_s}(i) \mathbf{x_K}(i)   
\end{equation}
The context vector $\mathbf{c_v}$ is analogous to the attention matrix $\mathbf{a}$ in \cref{eq:self_attn_transformer} in a sense that it also encodes the information from all tokens in the input $\mathbf{x}$, but is cheap to compute. 

The contextual information encoded in $\mathbf{c_v}$ is shared with all tokens in $\mathbf{x}$. To do so, the input $\mathbf{x}$ is linearly projected to a $d$-dimensional space using a value branch $\mathcal{V}$ with weights $\mathbf{W_V} \in \mathbb{R}^{d \times d}$, followed by a ReLU activation to produce an output $\mathbf{x_V} \in \mathbb{R}^{k \times d}$. The contextual information in $\mathbf{c_v}$ is then propagated to $\mathbf{x_V}$ via broadcasted element-wise multiplication operation. The resultant output is then fed to another linear layer with weights $\mathbf{W_O} \in \mathbb{R}^{d \times d}$ to produce the final output $\mathbf{y} \in \mathbb{R}^{k \times d}$. Mathematically, separable self-attention can be defined as:
\begin{equation}
\resizebox{0.5\hsize}{!}{
    $\mathbf{y} = \left( \underbrace{\sum \left( \overbrace{\sigma \left( \mathbf{x} \mathbf{W_I} \right)}^{\mathbf{c_s} \in \mathbb{R}^{k} } * \mathbf{x} \mathbf{W_K} \right)}_{\mathbf{c_v} \in \mathbb{R}^{d}}\ *\ \text{ReLU} \left(\mathbf{x} \mathbf{W_V} \right) \right) \mathbf{W_O}$
}
    \label{eq:context_vector_eq}
\end{equation}
where $*$ and $\sum$ are broadcastable element-wise multiplication and summation operations, respectively.

\paragraph{Comparison with self-attention methods} \cref{fig:compare_different_attn_layers} compares the proposed method with Transformer and Linformer. Because time complexity of self-attention methods do not account for the cost of operations that are used to implement these methods, some of the operations may become bottleneck on resource-constrained devices. For holistic understanding, module-level latency on a single CPU core with varying $k$ is also measured in addition to theoretical metrics. The proposed separable self-attention is fast and efficient as compared to MHA in Transformer and Linformer. 

Besides these module-level results, when we replaced the MHA in the transformer with the proposed self-separable attention in the MobileViT architecture, we observe $3 \times$ improvement in inference speed with similar performance on the ImageNet-1k dataset (\cref{tab:mobilevit_diff_attn}). These results show the efficacy of the proposed separable self-attention at the architecture-level. Note that self-attention in Transformer and Linformer yields similar results for MobileViT. This is because the number of tokens $k$ in MobileViT is fewer ($k \leq 1024$) as compared to language models, where Linformer is significantly faster than the transformer.

\paragraph{Relationship with additive addition} The proposed approach resembles the attention mechanism of \citet{bahdanau2014neural}, which also encodes the global information by taking a weighted-sum of LSTM outputs at each time step. Unlike \cite{bahdanau2014neural}, where input tokens interact via recurrence, the input tokens in the proposed method interact only with a latent token.

\begin{table}[t!]
    \centering
    \caption{\textbf{Effect of different self-attention methods} on the performance of MobileViT \cite{mehta2022mobilevit} on the ImageNet-1k dataset. Here, all models have similar number of parameters and FLOPs, and latency is measured on iPhone12.}
    \label{tab:mobilevit_diff_attn}
    \resizebox{0.55\columnwidth}{!}{
    \begin{tabular}{lrc}
        \toprule[1.5pt]
        \textbf{Attention unit} & \textbf{Latency} \contour{black}{$\downarrow$} & \textbf{Top-1} \contour{black}{$\uparrow$} \\
        \midrule[1.25pt]
         Self-attention in Transformer (\cref{fig:transformer_self_attn}; \cite{vaswani2017attention}) & 9.9 ms & \textbf{78.4} \\
         Self-attention in Linformer (\cref{fig:linformer_self_attn}; \cite{wang2020linformer}) & 10.2 ms & 78.2 \\
         Separable self-attention (Ours; \cref{fig:eff_self_attn}) & \textbf{3.4} ms & 78.1 \\
         \bottomrule[1.5pt]
    \end{tabular}
    }
\end{table}

\subsection{\arch architecture} 
\label{ssec:mobilevitv2_arch}
To demonstrate the effectiveness of the proposed separable self-attention on resource-constrained devices, we integrate separable self-attention with a recent ViT-based model, MobileViT \cite{mehta2022mobilevit}. MobileViT is a light-weight,  mobile-friendly hybrid network that delivers significantly better performance than other competitive CNN-based, transformer-based, or hybrid models, including MobileNets \cite{howard2017mobilenets, sandler2018mobilenetv2, howard2019searching}. To avoid ambiguity, we refer to MobileViT as MobileViTv1 in the rest of the paper. 

Specifically, we replace MHA in the transformer block in the MobileViTv1 with the proposed separable self-attention method. We call the resultant architecture \arch. We also do not use the skip-connection and fusion block in the MobileViT block (Fig. 1b in \cite{mehta2022mobilevit}) as it improves the performance marginally (Fig. 12 in \cite{mehta2022mobilevit}). Furthermore, to create \arch models at different complexities, we uniformly scale the width of \arch network using a width multiplier $\alpha \in \{0.5, 2.0\}$. This is in contrast to MobileViTv1 which trains three specific architectures (XXS, XS, and S) for mobile devices. More details about \arch's architecture are given in \cref{sec:app_architecture}.

\section{Experimental results}
\label{sec:exp_results}

\subsection{Object classification on the ImageNet dataset}
\label{ssec:imagenet_res}

\paragraph{Training on ImageNet-1k from scratch} We train \arch for 300 epochs with an effective batch size of 1024 images (128 images per GPU $\times$ 8 GPUs) using AdamW \cite{loshchilov2018decoupled} on the ImageNet-1k dataset \cite{russakovsky2015imagenet} with 1.28 million and 50 thousand training and validation images respectively. We linearly increase the learning rate from $10^{-6}$ to $0.002$ for the first 20k iterations. After that, the learning rate is decayed using a cosine annealing policy \cite{loshchilov2016sgdr}. To reduce stochastic noise during training, we use exponential moving average (EMA) \cite{polyak1992acceleration} as we find it helps larger models. We implement our models using \texttt{CVNets} \cite{mehta2022mobilevit, mehta2022cvnets}, and use their provided scripts for data processing, training, and evaluation.

\paragraph{Pre-training on ImageNet-21k-P and finetuning on ImageNet-1k} We train on the ImageNet-21k (winter'21 release) that contains about 13 million images across 19k classes. Specifically, we follow \cite{ridnik2021imagenet} to pre-process (e.g., remove classes with fewer samples) the dataset and split it into about 11 million and 522 thousand training and validation images spanning over 10,450 classes, respectively. Following \cite{ridnik2021imagenet}, we refer to this pre-processed dataset as ImageNet-21k-P. Note that the ImageNet-21k-P validation set does not overlap with the validation and test sets of ImageNet-1k.

We follow \cite{ridnik2021imagenet} for pre-training \arch on ImageNet-21k-P. For faster convergence, we initialize \arch models with ImageNet-1k weights and finetune it on ImageNet-21k-P for 80 epochs with an effective batch size of 4096 images (128 images per GPU x 32 GPUs). We do not use any linear warm-up. Other settings follow ImageNet-1k training.

We finetune ImageNet-21k-P pre-trained models on ImageNet-1k for 50 epochs using SGD with momentum (0.9) and cosine annealing policy with an effective batch size of 256 images (128 images per GPU $\times$ 2 GPUs).

\paragraph{Finetuning at higher resolution} \arch is a hybrid architecture that combines convolution and separable self-attention to learn visual representations. Unlike many ViT-based models (e.g., DeiT), \arch does not require adjustment to patch embeddings or positional biases for different input resolutions and is simple to finetune. We finetune \arch models at higher resolution (i.e., $384$$\times$$384$) for 10 epochs with a fixed learning rate of $10^{-3}$ using SGD.

\begin{table}[t!]
    \centering
    \caption{\textbf{Classification performance on the ImageNet-1k validation set}. Here, NS means that we are not able to measure the latency on mobile device as some operations (e.g., cyclic shifts) are not supported on mobile devices. Following \cite{mehta2022mobilevit}, latency is measured on iPhone12 with a batch size of 1. Similar to \cite{liu2021swin, liu2022convnet}, throughput is measured on NVIDIA V100 GPUs with a batch size of 128. The rows are grouped by network parameters.}
    \label{tab:compare_existing_methods}
    \resizebox{\columnwidth}{!}{
    \begin{tabular}{llccccccccc}
        \toprule[1.5pt]
        \multirow{2}{*}{\textbf{Row \#}} & \multirow{2}{*}{\textbf{Model}} & \multirow{2}{*}{\textbf{Type}} & \textbf{Neural} & \textbf{Extra} & \textbf{Image} & \multirow{2}{*}{\textbf{\# Params} \contour{black}{$\downarrow$}} & \multirow{2}{*}{\textbf{FLOPs} \contour{black}{$\downarrow$}} & \textbf{Latency} \contour{black}{$\downarrow$} & \textbf{Throughput} \contour{black}{$\uparrow$} & \textbf{Top-1} \contour{black}{$\uparrow$} \\
         & & & \textbf{search?} & \textbf{data} & \textbf{size} &  & & \textbf{(in ms)} & \textbf{(images/ sec)} & \textbf{(in \%)} \\
        \midrule[1.25pt]
         R1 & MobileViT-XXS \cite{mehta2022mobilevit} & Hybrid & \xmark & None & $256^2$ & \textbf{1.3 M} & 0.4 G & 4.8 & 4225 & 69.0 \\
         R2 & \arch-0.5 & Hybrid & \xmark & None & $256^2$ & 1.4 M & 0.5 G & \textbf{1.6} & \textbf{4595} & \textbf{70.2} \\
         R3 & MobileFormer-52 \cite{chen2021mobile} & Hybrid & \xmark & None & $224^2$ & 3.6 M & \textbf{52 M} & 7.1 & 4445 & 68.7 \\
         \midrule
         R4 & \arch-1.0 & Hybrid  & \xmark & None & $256^2$ & \textbf{4.9 M} & 1.8 G & 3.4 & 2351 & 78.1 \\
         R5 & EfficientNet-b0 \cite{tan2019efficientnet} & CNN & \cmark & None & $224^2$ & 5.3 M & \textbf{422 M} & \textbf{1.6} & \textbf{4619} & 77.1 \\
         R6 & DeiT-Tiny \cite{touvron2021training} & Transformer & \xmark & None & $224^2$ & 5.5 M & 1.3 G & 3.4 & 4541 & 72.2 \\
         R7 & MobileViT-S \cite{mehta2022mobilevit} & Hybrid  & \xmark & None & $256^2$ & 5.6 M & 2.0 G & 3.4 & 1986 & \textbf{78.4} \\
         \midrule
         R8 & EfficientNet-b2 \cite{tan2019efficientnet} & CNN & \cmark & None & $288^2$ & \textbf{9.1 M} & 1.2 G & \textbf{3.8} & \textbf{2032} & 80.1 \\
         R9 & \arch-1.5 &  Hybrid  & \xmark & None & $256^2$ & 10.6 M & 4.0 G & 5.1 & 1418 & \textbf{80.4} \\
         R10 & MobileFormer-294 \cite{chen2021mobile} & Hybrid & \xmark & None & $224^2$ & 11.8 M & \textbf{294 M} & 40.7 & 1402 & 77.9 \\
         \midrule
         R11 &  \arch-2.0 &  Hybrid & \xmark & None & $256^2$ & \textbf{18.5 M} & 7.5 G & 7.5 & 1105 & 81.2 \\
         R12 & Swin-T \cite{liu2021swin} & Hybrid & \xmark & None & $224^2$ & 28.3 M & \textbf{4.5 G} & NS & 1390 & 81.3 \\
         R13 & ConvNext-T \cite{liu2022convnet} & CNN  & \xmark & None & $224^2$ & 28.6 M & \textbf{4.5 G} & \textbf{3.7} & \textbf{1800} & 82.1 \\
         R14 & DeiT-Base \cite{touvron2021training} & Transformer & \xmark & None & $224^2$ & 86.6 M & 17.6 G & 13.2 & 958 & 81.8 \\
         \midrule
         R15 &  \arch-2.0 &  Hybrid & \xmark & ImageNet-21k-P & $256^2$ & \textbf{18.5 M} & 7.5 G & 7.5 & 1105 & 82.4 \\
         R16 & ConvNext-T \cite{liu2022convnet} & CNN & \xmark & ImageNet-21k & $224^2$ & 28.6 M & \textbf{4.5 G} & \textbf{3.7} & \textbf{1800} & \textbf{82.9} \\
        \midrule
        R17 &  \arch-2.0 &  Hybrid & \xmark & ImageNet-21k-P & $384^2$ & \textbf{18.5 M} & 16.1 G & 17.0 & 488 & 83.4 \\
        R18 & ConvNext-T \cite{liu2022convnet} & CNN & \xmark & ImageNet-21k & $384^2$ & 28.6 M & \textbf{13.1 G} & \textbf{8.6} & \textbf{645} & \textbf{84.1} \\
         \bottomrule[1.5pt]
    \end{tabular}
    }
\end{table}

\paragraph{Comparison with existing methods} \cref{tab:compare_existing_methods} and \cref{fig:compare_mobilevitv1_mobilevitv2} compares \arch's performance with recent methods\footnote{For additional results including ablations, see \cref{sec:app_mvitv2_in1k}, \cref{sec:app_mvitv2_light}, and \cref{sec:app_mvitv2_ablate}.}. We make following observations: 
\begin{itemize}[leftmargin=*]
    \item When MHA in MobileViTv1 is replaced with separable self-attention, the resultant model, \arch, is faster and better (\cref{fig:compare_mobilevitv1_mobilevitv2}); validating the effectiveness of the proposed separable self-attention method for mobile ViTs.
    \item Compared to transformer-based (including hybrid) models, \arch~models are fast on mobile devices. For example, \arch~is about $8 \times$ faster on a mobile device and delivers 2.5\% better performance on the ImageNet-1k dataset than MobileFormer \cite{chen2021mobile}, even though MobileFormer is FLOP efficient (R9 vs. R10). However, on GPU, both MobileFormer and \arch run at a similar speed. The discrepancy in FLOPs and speed of MobileFormer across devices is primarily because of its architectural design. MobileFormer has conditional operations between mobile and former blocks. Such conditional operations, especially on resource-constrained devices, have a low degree of parallelism and create memory bottlenecks, resulting in a high latency network. \citet{ma2018shufflenet} also makes a similar observation for CNN-based architectures.
    \item \arch~bridges the latency gap between CNN- and ViT-based models on mobile devices while maintaining performance with similar or fewer parameters. For example, on a mobile device, ConvNexT \cite{liu2022convnet} (CNN-based model) is $2\times$ and $3.6 \times$ faster than \arch (hybrid model) and DeiT (transformer-based model) for similar performance respectively (see R11, R13, and R14). The low latency of fully CNN-based models on mobile devices can be attributed to several device-level optimizations that have been done for CNN-based models over the past few years (e.g., dedicated hardware implementations for convolutions and folding batch normalization with convolutions). ViT-based models still lack such optimizations and therefore, the resultant inference graphs are sub-optimal. Though \arch~bridges the latency gap between CNNs and ViTs, we believe the latency of ViT-based models will improve in the future with similar optimizations.
    \item The delta in speed (on GPU) between ConvNext and \arch (R15-R18) at higher model complexities reduces from $1.6\times$ to $1.3 \times$ when input resolution is increased from $224 \times 224$ (or $256 \times 256$) to $384 \times 384$, suggesting ViT-based (including hybrid) models exhibit better scaling properties as compared to CNNs. This is because of a higher degree of parallelism that ViT-based models offer at a large scale \cite{dosovitskiy2021an, brown2020language}. Our results on down-stream tasks in \cref{ssec:down_stream_tasks} and previous work on scaling ViTs \cite{dosovitskiy2021an,zhai2021scaling} further supports this observation.
\end{itemize}

\subsection{Evaluation on down-stream tasks}
\label{ssec:down_stream_tasks}

\paragraph{Semantic segmentation} We integrate \arch~with two standard segmentation architectures, PSPNet \cite{zhao2017pyramid} and DeepLabv3 \cite{chen2017rethinking}, and study it on two standard semantic segmentation datasets, ADE20k \cite{zhou2017scene} and PASCAL VOC 2012 \cite{everingham2015pascal}. For training details including hyper-parameters, see supplementary material. 
 
\cref{tab:sem_seg_res} and \cref{fig:compre_mv1_mv2_deeplab} compares the segmentation performance in terms of validation mean intersection over union (mIOU) of \arch with different segmentation methods. \arch~delivers competitive performance at different complexities while having significantly fewer parameters and FLOPs. Interestingly, the inference speed of \arch~models is comparable to CNN-based models, including light-weight MobileNetv2 and heavy-weight ResNet-50 \cite{he2016deep} model. This is consistent with our observation in \cref{ssec:imagenet_res} (R17 vs. R18; \cref{tab:compare_existing_methods}) where we also observe that ViT-based models scale better than CNN's at higher input resolutions and model complexities.

\begin{table}[t!]
    \centering
    \caption{\textbf{Semantic segmentation results on the ADE20k and the PASCAL VOC 2012 datasets.} Here, throughput, network parameters, and FLOPs are measured on the ADE20k dataset for an input with a spatial resolution of $512 \times 512$. mIoU (mean intersection over union) score is calculated for a single scale only. Throughput is calculated using a batch size of 32 images on a single NVIDIA V100 GPU with 32 GB memory and is an average of over 50 iterations (excluding 10 iterations for warmup). We do not report latency on a mobile device as some of the operations (e.g., pyramid pooling in PSPNet) are not optimally implemented for mobile devices. The baseline results are from the MMSegmentation library \cite{mmseg2020}. Rows are grouped by network parameters.}
    \label{tab:sem_seg_res}
    \resizebox{\columnwidth}{!}{
    \begin{tabular}{ccccrrcc}
        \toprule[1.5pt]
        \textbf{Seg.} & \textbf{ImageNet-1k} & \textbf{Image} & \textbf{Throughput} \contour{black}{$\uparrow$} & \multicolumn{1}{c}{\textbf{\# Params} \contour{black}{$\downarrow$}} & \multicolumn{1}{c}{\textbf{FLOPs} \contour{black}{$\downarrow$}} & \multicolumn{2}{c}{\textbf{mIoU} \contour{black}{$\uparrow$}} \\
        \cmidrule[1.25pt]{7-8}
        \textbf{Model} & \textbf{Backbone} & \textbf{Size} & \textbf{(images/sec)} & \textbf{(in millions)} & \textbf{(in billions)} & \textbf{ADE20k} \cite{zhou2017scene} & \textbf{PASCAL VOC} \cite{everingham2015pascal} \\
        \midrule[1.25pt]
        \multirow{2}{*}{PSPNet \cite{zhao2017pyramid}} & \arch-0.5 (Ours) & $512^2$ & \textbf{439} & \textbf{3.6 M} & \textbf{15.4 G} & \textbf{31.8} & 74.6 \\
          & MobileNetv2 \cite{sandler2018mobilenetv2} & $512^2$ & 276  & 13.7 M & 53.1 G & 29.7 & -- \\
         \hdashline
         \multirow{2}{*}{PSPNet \cite{zhao2017pyramid}} & \arch-1.75 (Ours) & $512^2$ & 114 & \textbf{22.5 M} & \textbf{95.9 G} & 39.8 & \textbf{80.2} \\
          & ResNet-50 \cite{he2016deep} & $512^2$ & \textbf{119} & 49.1 M & 179.1 G & \textbf{41.1} & 76.8 \\
         \midrule
         \multirow{2}{*}{DeepLabv3 \cite{chen2017rethinking}} & \arch-0.75 (Ours) & $512^2$ & 241 & \textbf{9.6 M} & \textbf{40.0 G} & \textbf{34.7} & 75.1 ($\alpha$=0.5) \\ 
          & MobileNetv2 \cite{sandler2018mobilenetv2} & $512^2$ & \textbf{246} & 18.7 M & 75.4 G & 34.1 & -- \\
         \hdashline
         \multirow{2}{*}{DeepLabv3 \cite{chen2017rethinking}} & \arch-2.0 (Ours) & $512^2$ & 90 & \textbf{34.0 M} & \textbf{147.0 G} & 40.9 & \textbf{80.3} ($\alpha$=1.5) \\ 
         & ResNet-50 \cite{he2016deep} & $512^2$ & \textbf{103} & 68.2 M & 270.3 G & \textbf{42.4} & 79.1 \\
         \bottomrule[1.5pt]
    \end{tabular}
    }
\end{table}

\paragraph{Object detection} We integrate \arch~with SSDLite \cite{sandler2018mobilenetv2} (SSD head \cite{liu2016ssd} with separable convolutions) for mobile object detection, and study its performance on MS-COCO dataset \cite{lin2014microsoft}. We follow \cite{mehta2022mobilevit} for training detection models. \cref{tab:obj_det_res} and \cref{fig:compre_mv1_mv2_ssd} compares SSDLite's detection performance in terms of validation mean average precision (mAP) using different ImageNet-1k backbones. \arch~delivers competitive performance to models with different capacities, further validating the effectiveness of the proposed self-separable attention method.

\begin{table}[t!]
    \centering
    \caption{\textbf{Object detection using SSDLite on the MS-COCO dataset.} Here, throughput is measured with a batch of 128 images on the NVIDIA V100 GPU, and is an average over 50 iterations (excluding 10 iterations for warmup). Latency on a mobile device is not reported as some operations (e.g., hard swish) are not optimally implemented for such devices. Rows are grouped by network parameters.}
    \label{tab:obj_det_res}
    \resizebox{0.8\columnwidth}{!}{
    \begin{tabular}{lcccccc}
        \toprule[1.5pt]
        \multicolumn{1}{c}{\textbf{ImageNet-1k}} & \textbf{Image} & \textbf{Throughput} \contour{black}{$\uparrow$} & \textbf{\# Params} \contour{black}{$\downarrow$} & \textbf{FLOPs} \contour{black}{$\downarrow$} & \multirow{2}{*}{\textbf{mAP} \contour{black}{$\uparrow$}} \\
        \multicolumn{1}{c}{\textbf{backbone}} & \textbf{Size} & \textbf{(images/sec)} & \textbf{(in millions)} & \textbf{(in billions)} &  \\
        \midrule[1.25pt]
        MobileViTv1-XXS & $320^2$ & 2246 & \textbf{1.7 M} & 0.9 G & 19.9 \\
        \arch-0.5 (Ours) & $320^2$ & 2782 & 2.0 M & 0.9 G & 21.2 \\
        \arch-0.75 (Ours) & $320^2$ & 1876 & 3.6 M & 1.8 G & \textbf{24.6} \\
        Mobilenetv2 & $320^2$ & 3052 & 4.3 M & 0.8 G & 22.1 \\
        MobileNetv3 & $320^2$ & 3884 & 5.0 M & \textbf{0.6 G} & 22.0 \\
        MobileNetv1 & $320^2$ & \textbf{4330} & 5.1 M & 1.3 G & 22.2 \\
        \midrule
        \arch-1.75 & $320^2$ & 780 & \textbf{14.9 M} & \textbf{9.0 G} & \textbf{29.5} \\
        ResNet-50 & $300^2$ & 744 & 22.9 M & 20.2 G & 25.2 \\
         \bottomrule[1.5pt]
    \end{tabular}
    }
\end{table}

\section{Visualizations of self-separable attention scores}
\begin{figure}[t!]
    \centering
   %\resizebox{\columnwidth}{!}{
    \begin{tabular}{c}
         \includegraphics[height=70px, width=70px]{attn_map_vis/sample_5/000000134112.jpg} \includegraphics[height=70px]{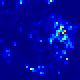} \includegraphics[height=70px]{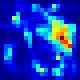} \includegraphics[height=70px]{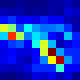}  \\
          \includegraphics[height=70px, width=70px]{attn_map_vis/sample_6/000000105264.jpg} \includegraphics[height=70px]{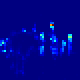} \includegraphics[height=70px]{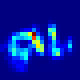} \includegraphics[height=70px]{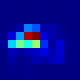}
    \end{tabular}
   %}
    \caption{\textbf{Context score maps at different output strides (OS) of \arch model.} Observe how context scores pay attention to semantically relevant image regions. (\textbf{Left to right:} input image, context scores at OS=8, context scores at OS=16, and context scores at OS=32). For more examples and details about context score map generation, see \cref{sec:app_vis}.}
    \label{fig:self_sep_visualizations}
\end{figure}

\cref{fig:self_sep_visualizations} visualizes what the context scores learn at different output strides\footnote{Output stride is the ratio of the spatial dimension of the input to the feature map.} of \arch network. We found that separable self-attention layers pay attention to low-, mid-, and high-level features, and allow \arch to learn representations from semantically relevant image regions. 

\section{Conclusions}
Transformer-based vision models are slow on mobile devices as compared to CNN-based models because multi-headed self-attention is expensive on resource-constrained devices. In this paper, we introduce a \emph{separable self-attention} method that has linear complexity and can be implemented using hardware-friendly element-wise operations. Experimental results on standard datasets and tasks demonstrate the effectiveness of the proposed method over multi-headed self-attention.  

\section*{Acknowledgements}
We are grateful to Ali Farhadi, Peter Zatloukal, Oncel Tuzel, Rick Chang, Fartash Faghri, Farzad Abdolhosseini, Lailin Chen, and Max Horton for their helpful comments. We are also thankful to Apple's infrastructure and open-source teams for their help with training infrastructure and open-source release of the code and pre-trained models.

\small{
\bibliographystyle{unsrtnat}
\bibliography{main}

\begin{thebibliography}{59}
\providecommand{\natexlab}[1]{#1}
\providecommand{\url}[1]{\texttt{#1}}
\expandafter\ifx\csname urlstyle\endcsname\relax
  \providecommand{\doi}[1]{doi: #1}\else
  \providecommand{\doi}{doi: \begingroup \urlstyle{rm}\Url}\fi

\bibitem[Dosovitskiy et~al.(2021)Dosovitskiy, Beyer, Kolesnikov, Weissenborn,
  Zhai, Unterthiner, Dehghani, Minderer, Heigold, Gelly, Uszkoreit, and
  Houlsby]{dosovitskiy2021an}
Alexey Dosovitskiy, Lucas Beyer, Alexander Kolesnikov, Dirk Weissenborn,
  Xiaohua Zhai, Thomas Unterthiner, Mostafa Dehghani, Matthias Minderer, Georg
  Heigold, Sylvain Gelly, Jakob Uszkoreit, and Neil Houlsby.
\newblock An image is worth 16x16 words: Transformers for image recognition at
  scale.
\newblock In \emph{International Conference on Learning Representations}, 2021.
\newblock URL \url{https://openreview.net/forum?id=YicbFdNTTy}.

\bibitem[Touvron et~al.(2021)Touvron, Cord, Douze, Massa, Sablayrolles, and
  J{\'e}gou]{touvron2021training}
Hugo Touvron, Matthieu Cord, Matthijs Douze, Francisco Massa, Alexandre
  Sablayrolles, and Herv{\'e} J{\'e}gou.
\newblock Training data-efficient image transformers \& distillation through
  attention.
\newblock In \emph{International Conference on Machine Learning}, pages
  10347--10357. PMLR, 2021.

\bibitem[Liu et~al.(2021)Liu, Lin, Cao, Hu, Wei, Zhang, Lin, and
  Guo]{liu2021swin}
Ze~Liu, Yutong Lin, Yue Cao, Han Hu, Yixuan Wei, Zheng Zhang, Stephen Lin, and
  Baining Guo.
\newblock Swin transformer: Hierarchical vision transformer using shifted
  windows.
\newblock In \emph{Proceedings of the IEEE/CVF International Conference on
  Computer Vision}, pages 10012--10022, 2021.

\bibitem[Mehta and Rastegari(2022)]{mehta2022mobilevit}
Sachin Mehta and Mohammad Rastegari.
\newblock Mobilevit: Light-weight, general-purpose, and mobile-friendly vision
  transformer.
\newblock In \emph{International Conference on Learning Representations}, 2022.
\newblock URL \url{https://openreview.net/forum?id=vh-0sUt8HlG}.

\bibitem[Vaswani et~al.(2017)Vaswani, Shazeer, Parmar, Uszkoreit, Jones, Gomez,
  Kaiser, and Polosukhin]{vaswani2017attention}
Ashish Vaswani, Noam Shazeer, Niki Parmar, Jakob Uszkoreit, Llion Jones,
  Aidan~N Gomez, {\L}ukasz Kaiser, and Illia Polosukhin.
\newblock Attention is all you need.
\newblock \emph{Advances in neural information processing systems}, 30, 2017.

\bibitem[Paszke et~al.(2019)Paszke, Gross, Massa, Lerer, Bradbury, Chanan,
  Killeen, Lin, Gimelshein, Antiga, et~al.]{paszke2019pytorch}
Adam Paszke, Sam Gross, Francisco Massa, Adam Lerer, James Bradbury, Gregory
  Chanan, Trevor Killeen, Zeming Lin, Natalia Gimelshein, Luca Antiga, et~al.
\newblock Pytorch: An imperative style, high-performance deep learning library.
\newblock \emph{Advances in neural information processing systems}, 32, 2019.

\bibitem[Child et~al.(2019)Child, Gray, Radford, and
  Sutskever]{child2019generating}
Rewon Child, Scott Gray, Alec Radford, and Ilya Sutskever.
\newblock Generating long sequences with sparse transformers.
\newblock \emph{arXiv preprint arXiv:1904.10509}, 2019.

\bibitem[Kitaev et~al.(2020)Kitaev, Kaiser, and Levskaya]{kitaev2020reformer}
Nikita Kitaev, {\L}ukasz Kaiser, and Anselm Levskaya.
\newblock Reformer: The efficient transformer.
\newblock \emph{arXiv preprint arXiv:2001.04451}, 2020.

\bibitem[Beltagy et~al.(2020)Beltagy, Peters, and Cohan]{beltagy2020longformer}
Iz~Beltagy, Matthew~E Peters, and Arman Cohan.
\newblock Longformer: The long-document transformer.
\newblock \emph{arXiv preprint arXiv:2004.05150}, 2020.

\bibitem[Wang et~al.(2020)Wang, Li, Khabsa, Fang, and Ma]{wang2020linformer}
Sinong Wang, Belinda~Z Li, Madian Khabsa, Han Fang, and Hao Ma.
\newblock Linformer: Self-attention with linear complexity.
\newblock \emph{arXiv preprint arXiv:2006.04768}, 2020.

\bibitem[Parmar et~al.(2018)Parmar, Vaswani, Uszkoreit, Kaiser, Shazeer, Ku,
  and Tran]{parmar2018image}
Niki Parmar, Ashish Vaswani, Jakob Uszkoreit, Lukasz Kaiser, Noam Shazeer,
  Alexander Ku, and Dustin Tran.
\newblock Image transformer.
\newblock In \emph{International Conference on Machine Learning}, pages
  4055--4064. PMLR, 2018.

\bibitem[Qiu et~al.(2019)Qiu, Ma, Levy, Yih, Wang, and Tang]{qiu2019blockwise}
Jiezhong Qiu, Hao Ma, Omer Levy, Scott Wen-tau Yih, Sinong Wang, and Jie Tang.
\newblock Blockwise self-attention for long document understanding.
\newblock \emph{arXiv preprint arXiv:1911.02972}, 2019.

\bibitem[Vyas et~al.(2020)Vyas, Katharopoulos, and Fleuret]{vyas2020fast}
Apoorv Vyas, Angelos Katharopoulos, and Fran{\c{c}}ois Fleuret.
\newblock Fast transformers with clustered attention.
\newblock \emph{Advances in Neural Information Processing Systems},
  33:\penalty0 21665--21674, 2020.

\bibitem[Wang et~al.(2021{\natexlab{a}})Wang, Zhou, Gan, Chen, Fang, Sun,
  Cheng, and Liu]{wang2021cluster}
Shuohang Wang, Luowei Zhou, Zhe Gan, Yen-Chun Chen, Yuwei Fang, Siqi Sun,
  Yu~Cheng, and Jingjing Liu.
\newblock Cluster-former: Clustering-based sparse transformer for question
  answering.
\newblock In \emph{Findings of the Association for Computational Linguistics:
  ACL-IJCNLP 2021}, pages 3958--3968, 2021{\natexlab{a}}.

\bibitem[Choromanski et~al.(2020)Choromanski, Likhosherstov, Dohan, Song, Gane,
  Sarlos, Hawkins, Davis, Mohiuddin, Kaiser, et~al.]{choromanski2020rethinking}
Krzysztof Choromanski, Valerii Likhosherstov, David Dohan, Xingyou Song,
  Andreea Gane, Tamas Sarlos, Peter Hawkins, Jared Davis, Afroz Mohiuddin,
  Lukasz Kaiser, et~al.
\newblock Rethinking attention with performers.
\newblock \emph{arXiv preprint arXiv:2009.14794}, 2020.

\bibitem[Mehta et~al.(2021)Mehta, Ghazvininejad, Iyer, Zettlemoyer, and
  Hajishirzi]{mehta2021delight}
Sachin Mehta, Marjan Ghazvininejad, Srinivasan Iyer, Luke Zettlemoyer, and
  Hannaneh Hajishirzi.
\newblock Delight: Deep and light-weight transformer.
\newblock In \emph{International Conference on Learning Representations}, 2021.
\newblock URL \url{https://openreview.net/forum?id=ujmgfuxSLrO}.

\bibitem[Wu et~al.(2021)Wu, Xiao, Codella, Liu, Dai, Yuan, and
  Zhang]{wu2021cvt}
Haiping Wu, Bin Xiao, Noel Codella, Mengchen Liu, Xiyang Dai, Lu~Yuan, and Lei
  Zhang.
\newblock Cvt: Introducing convolutions to vision transformers.
\newblock In \emph{Proceedings of the IEEE/CVF International Conference on
  Computer Vision}, pages 22--31, 2021.

\bibitem[Heo et~al.(2021)Heo, Yun, Han, Chun, Choe, and Oh]{heo2021rethinking}
Byeongho Heo, Sangdoo Yun, Dongyoon Han, Sanghyuk Chun, Junsuk Choe, and
  Seong~Joon Oh.
\newblock Rethinking spatial dimensions of vision transformers.
\newblock In \emph{Proceedings of the IEEE/CVF International Conference on
  Computer Vision}, pages 11936--11945, 2021.

\bibitem[Ryoo et~al.(2021)Ryoo, Piergiovanni, Arnab, Dehghani, and
  Angelova]{ryoo2021tokenlearner}
Michael Ryoo, AJ~Piergiovanni, Anurag Arnab, Mostafa Dehghani, and Anelia
  Angelova.
\newblock Tokenlearner: Adaptive space-time tokenization for videos.
\newblock \emph{Advances in Neural Information Processing Systems}, 34, 2021.

\bibitem[Wang et~al.(2021{\natexlab{b}})Wang, Xie, Li, Fan, Song, Liang, Lu,
  Luo, and Shao]{wang2021pyramid}
Wenhai Wang, Enze Xie, Xiang Li, Deng-Ping Fan, Kaitao Song, Ding Liang, Tong
  Lu, Ping Luo, and Ling Shao.
\newblock Pyramid vision transformer: A versatile backbone for dense prediction
  without convolutions.
\newblock In \emph{Proceedings of the IEEE/CVF International Conference on
  Computer Vision}, pages 568--578, 2021{\natexlab{b}}.

\bibitem[Micikevicius et~al.(2018)Micikevicius, Narang, Alben, Diamos, Elsen,
  Garcia, Ginsburg, Houston, Kuchaiev, Venkatesh, and
  Wu]{micikevicius2018mixed}
Paulius Micikevicius, Sharan Narang, Jonah Alben, Gregory Diamos, Erich Elsen,
  David Garcia, Boris Ginsburg, Michael Houston, Oleksii Kuchaiev, Ganesh
  Venkatesh, and Hao Wu.
\newblock Mixed precision training.
\newblock In \emph{International Conference on Learning Representations}, 2018.
\newblock URL \url{https://openreview.net/forum?id=r1gs9JgRZ}.

\bibitem[Dettmers et~al.(2022)Dettmers, Lewis, Shleifer, and
  Zettlemoyer]{dettmers2022bit}
Tim Dettmers, Mike Lewis, Sam Shleifer, and Luke Zettlemoyer.
\newblock 8-bit optimizers via block-wise quantization.
\newblock In \emph{International Conference on Learning Representations}, 2022.
\newblock URL \url{https://openreview.net/forum?id=shpkpVXzo3h}.

\bibitem[Zhai et~al.(2021{\natexlab{a}})Zhai, Kolesnikov, Houlsby, and
  Beyer]{xiaohua2021scaling}
Xiaohua Zhai, Alexander Kolesnikov, Neil Houlsby, and Lucas Beyer.
\newblock Scaling vision transformers, 2021{\natexlab{a}}.
\newblock URL \url{https://arxiv.org/abs/2106.04560}.

\bibitem[Sandler et~al.(2018)Sandler, Howard, Zhu, Zhmoginov, and
  Chen]{sandler2018mobilenetv2}
Mark Sandler, Andrew Howard, Menglong Zhu, Andrey Zhmoginov, and Liang-Chieh
  Chen.
\newblock Mobilenetv2: Inverted residuals and linear bottlenecks.
\newblock In \emph{Proceedings of the IEEE conference on computer vision and
  pattern recognition}, pages 4510--4520, 2018.

\bibitem[Howard et~al.(2019)Howard, Sandler, Chu, Chen, Chen, Tan, Wang, Zhu,
  Pang, Vasudevan, et~al.]{howard2019searching}
Andrew Howard, Mark Sandler, Grace Chu, Liang-Chieh Chen, Bo~Chen, Mingxing
  Tan, Weijun Wang, Yukun Zhu, Ruoming Pang, Vijay Vasudevan, et~al.
\newblock Searching for mobilenetv3.
\newblock In \emph{Proceedings of the IEEE/CVF International Conference on
  Computer Vision}, pages 1314--1324, 2019.

\bibitem[Bahdanau et~al.(2014)Bahdanau, Cho, and Bengio]{bahdanau2014neural}
Dzmitry Bahdanau, Kyunghyun Cho, and Yoshua Bengio.
\newblock Neural machine translation by jointly learning to align and
  translate.
\newblock \emph{arXiv preprint arXiv:1409.0473}, 2014.

\bibitem[Howard et~al.(2017)Howard, Zhu, Chen, Kalenichenko, Wang, Weyand,
  Andreetto, and Adam]{howard2017mobilenets}
Andrew~G Howard, Menglong Zhu, Bo~Chen, Dmitry Kalenichenko, Weijun Wang,
  Tobias Weyand, Marco Andreetto, and Hartwig Adam.
\newblock Mobilenets: Efficient convolutional neural networks for mobile vision
  applications.
\newblock \emph{arXiv preprint arXiv:1704.04861}, 2017.

\bibitem[Loshchilov and Hutter(2019)]{loshchilov2018decoupled}
Ilya Loshchilov and Frank Hutter.
\newblock Decoupled weight decay regularization.
\newblock In \emph{International Conference on Learning Representations}, 2019.
\newblock URL \url{https://openreview.net/forum?id=Bkg6RiCqY7}.

\bibitem[Russakovsky et~al.(2015)Russakovsky, Deng, Su, Krause, Satheesh, Ma,
  Huang, Karpathy, Khosla, Bernstein, et~al.]{russakovsky2015imagenet}
Olga Russakovsky, Jia Deng, Hao Su, Jonathan Krause, Sanjeev Satheesh, Sean Ma,
  Zhiheng Huang, Andrej Karpathy, Aditya Khosla, Michael Bernstein, et~al.
\newblock Imagenet large scale visual recognition challenge.
\newblock \emph{International journal of computer vision}, 115\penalty0
  (3):\penalty0 211--252, 2015.

\bibitem[Loshchilov and Hutter(2017)]{loshchilov2016sgdr}
Ilya Loshchilov and Frank Hutter.
\newblock Sgdr: Stochastic gradient descent with warm restarts.
\newblock In \emph{International Conference on Learning Representations}, 2017.

\bibitem[Polyak and Juditsky(1992)]{polyak1992acceleration}
Boris~T Polyak and Anatoli~B Juditsky.
\newblock Acceleration of stochastic approximation by averaging.
\newblock \emph{SIAM journal on control and optimization}, 30\penalty0
  (4):\penalty0 838--855, 1992.

\bibitem[Mehta et~al.(2022)Mehta, Abdolhosseini, and
  Rastegari]{mehta2022cvnets}
Sachin Mehta, Farzad Abdolhosseini, and Mohammad Rastegari.
\newblock Cvnets: High performance library for computer vision.
\newblock \emph{CoRR}, 2022.

\bibitem[Ridnik et~al.(2021)Ridnik, Ben-Baruch, Noy, and
  Zelnik-Manor]{ridnik2021imagenet}
Tal Ridnik, Emanuel Ben-Baruch, Asaf Noy, and Lihi Zelnik-Manor.
\newblock Imagenet-21k pretraining for the masses.
\newblock \emph{arXiv preprint arXiv:2104.10972}, 2021.

\bibitem[Liu et~al.(2022)Liu, Mao, Wu, Feichtenhofer, Darrell, and
  Xie]{liu2022convnet}
Zhuang Liu, Hanzi Mao, Chao-Yuan Wu, Christoph Feichtenhofer, Trevor Darrell,
  and Saining Xie.
\newblock A convnet for the 2020s.
\newblock \emph{arXiv preprint arXiv:2201.03545}, 2022.

\bibitem[Chen et~al.(2021{\natexlab{a}})Chen, Dai, Chen, Liu, Dong, Yuan, and
  Liu]{chen2021mobile}
Yinpeng Chen, Xiyang Dai, Dongdong Chen, Mengchen Liu, Xiaoyi Dong, Lu~Yuan,
  and Zicheng Liu.
\newblock Mobile-former: Bridging mobilenet and transformer.
\newblock \emph{arXiv preprint arXiv:2108.05895}, 2021{\natexlab{a}}.

\bibitem[Tan and Le(2019)]{tan2019efficientnet}
Mingxing Tan and Quoc Le.
\newblock Efficientnet: Rethinking model scaling for convolutional neural
  networks.
\newblock In \emph{International conference on machine learning}, pages
  6105--6114. PMLR, 2019.

\bibitem[Ma et~al.(2018)Ma, Zhang, Zheng, and Sun]{ma2018shufflenet}
Ningning Ma, Xiangyu Zhang, Hai-Tao Zheng, and Jian Sun.
\newblock Shufflenet v2: Practical guidelines for efficient cnn architecture
  design.
\newblock In \emph{Proceedings of the European conference on computer vision
  (ECCV)}, pages 116--131, 2018.

\bibitem[Brown et~al.(2020)Brown, Mann, Ryder, Subbiah, Kaplan, Dhariwal,
  Neelakantan, Shyam, Sastry, Askell, et~al.]{brown2020language}
Tom Brown, Benjamin Mann, Nick Ryder, Melanie Subbiah, Jared~D Kaplan, Prafulla
  Dhariwal, Arvind Neelakantan, Pranav Shyam, Girish Sastry, Amanda Askell,
  et~al.
\newblock Language models are few-shot learners.
\newblock \emph{Advances in neural information processing systems},
  33:\penalty0 1877--1901, 2020.

\bibitem[Zhai et~al.(2021{\natexlab{b}})Zhai, Kolesnikov, Houlsby, and
  Beyer]{zhai2021scaling}
Xiaohua Zhai, Alexander Kolesnikov, Neil Houlsby, and Lucas Beyer.
\newblock Scaling vision transformers.
\newblock \emph{CoRR}, abs/2106.04560, 2021{\natexlab{b}}.

\bibitem[Zhao et~al.(2017)Zhao, Shi, Qi, Wang, and Jia]{zhao2017pyramid}
Hengshuang Zhao, Jianping Shi, Xiaojuan Qi, Xiaogang Wang, and Jiaya Jia.
\newblock Pyramid scene parsing network.
\newblock In \emph{Proceedings of the IEEE conference on computer vision and
  pattern recognition}, pages 2881--2890, 2017.

\bibitem[Chen et~al.(2017)Chen, Papandreou, Schroff, and
  Adam]{chen2017rethinking}
Liang-Chieh Chen, George Papandreou, Florian Schroff, and Hartwig Adam.
\newblock Rethinking atrous convolution for semantic image segmentation.
\newblock \emph{arXiv preprint arXiv:1706.05587}, 2017.

\bibitem[Zhou et~al.(2017)Zhou, Zhao, Puig, Fidler, Barriuso, and
  Torralba]{zhou2017scene}
Bolei Zhou, Hang Zhao, Xavier Puig, Sanja Fidler, Adela Barriuso, and Antonio
  Torralba.
\newblock Scene parsing through ade20k dataset.
\newblock In \emph{Proceedings of the IEEE conference on computer vision and
  pattern recognition}, pages 633--641, 2017.

\bibitem[Everingham et~al.(2015)Everingham, Eslami, Van~Gool, Williams, Winn,
  and Zisserman]{everingham2015pascal}
Mark Everingham, SM~Eslami, Luc Van~Gool, Christopher~KI Williams, John Winn,
  and Andrew Zisserman.
\newblock The pascal visual object classes challenge: A retrospective.
\newblock \emph{International journal of computer vision}, 111\penalty0
  (1):\penalty0 98--136, 2015.

\bibitem[He et~al.(2016)He, Zhang, Ren, and Sun]{he2016deep}
Kaiming He, Xiangyu Zhang, Shaoqing Ren, and Jian Sun.
\newblock Deep residual learning for image recognition.
\newblock In \emph{Proceedings of the IEEE conference on computer vision and
  pattern recognition}, pages 770--778, 2016.

\bibitem[Contributors(2020)]{mmseg2020}
MMSegmentation Contributors.
\newblock {MMSegmentation}: Openmmlab semantic segmentation toolbox and
  benchmark.
\newblock \url{https://github.com/open-mmlab/mmsegmentation}, 2020.

\bibitem[Liu et~al.(2016)Liu, Anguelov, Erhan, Szegedy, Reed, Fu, and
  Berg]{liu2016ssd}
Wei Liu, Dragomir Anguelov, Dumitru Erhan, Christian Szegedy, Scott Reed,
  Cheng-Yang Fu, and Alexander~C Berg.
\newblock Ssd: Single shot multibox detector.
\newblock In \emph{European conference on computer vision}, pages 21--37.
  Springer, 2016.

\bibitem[Lin et~al.(2014)Lin, Maire, Belongie, Hays, Perona, Ramanan,
  Doll{\'a}r, and Zitnick]{lin2014microsoft}
Tsung-Yi Lin, Michael Maire, Serge Belongie, James Hays, Pietro Perona, Deva
  Ramanan, Piotr Doll{\'a}r, and C~Lawrence Zitnick.
\newblock Microsoft coco: Common objects in context.
\newblock In \emph{European conference on computer vision}, pages 740--755.
  Springer, 2014.

\bibitem[Elfwing et~al.(2018)Elfwing, Uchibe, and Doya]{elfwing2018sigmoid}
Stefan Elfwing, Eiji Uchibe, and Kenji Doya.
\newblock Sigmoid-weighted linear units for neural network function
  approximation in reinforcement learning.
\newblock \emph{Neural Networks}, 107:\penalty0 3--11, 2018.

\bibitem[Mehta et~al.(2019)Mehta, Rastegari, Shapiro, and
  Hajishirzi]{mehta2019espnetv2}
Sachin Mehta, Mohammad Rastegari, Linda Shapiro, and Hannaneh Hajishirzi.
\newblock Espnetv2: A light-weight, power efficient, and general purpose
  convolutional neural network.
\newblock In \emph{Proceedings of the IEEE/CVF Conference on Computer Vision
  and Pattern Recognition}, pages 9190--9200, 2019.

\bibitem[Yuan et~al.(2021)Yuan, Chen, Wang, Yu, Shi, Jiang, Tay, Feng, and
  Yan]{yuan2021tokens}
Li~Yuan, Yunpeng Chen, Tao Wang, Weihao Yu, Yujun Shi, Zihang Jiang, Francis~EH
  Tay, Jiashi Feng, and Shuicheng Yan.
\newblock Tokens-to-token vit: Training vision transformers from scratch on
  imagenet.
\newblock In \emph{Proceedings of the IEEE/CVF international conference on
  computer vision}, 2021.

\bibitem[Chen et~al.(2021{\natexlab{b}})Chen, Fan, and Panda]{chen2021crossvit}
Chun-Fu Chen, Quanfu Fan, and Rameswar Panda.
\newblock {CrossVit: Cross-attention multi-scale vision transformer for image
  classification}.
\newblock In \emph{Proceedings of the IEEE/CVF International Conference on
  Computer Vision (ICCV)}, 2021{\natexlab{b}}.

\bibitem[Li et~al.(2021)Li, Zhang, Cao, Timofte, and Van~Gool]{li2021localvit}
Yawei Li, Kai Zhang, Jiezhang Cao, Radu Timofte, and Luc Van~Gool.
\newblock Localvit: Bringing locality to vision transformers.
\newblock \emph{arXiv preprint arXiv:2104.05707}, 2021.

\bibitem[d'Ascoli et~al.(2021)d'Ascoli, Touvron, Leavitt, Morcos, Biroli, and
  Sagun]{d2021convit}
St{\'e}phane d'Ascoli, Hugo Touvron, Matthew Leavitt, Ari Morcos, Giulio
  Biroli, and Levent Sagun.
\newblock Convit: Improving vision transformers with soft convolutional
  inductive biases.
\newblock \emph{arXiv preprint arXiv:2103.10697}, 2021.

\bibitem[Szegedy et~al.(2015)Szegedy, Liu, Jia, Sermanet, Reed, Anguelov,
  Erhan, Vanhoucke, and Rabinovich]{szegedy2015going}
Christian Szegedy, Wei Liu, Yangqing Jia, Pierre Sermanet, Scott Reed, Dragomir
  Anguelov, Dumitru Erhan, Vincent Vanhoucke, and Andrew Rabinovich.
\newblock Going deeper with convolutions.
\newblock In \emph{Proceedings of the IEEE conference on computer vision and
  pattern recognition}, pages 1--9, 2015.

\bibitem[Cubuk et~al.(2020)Cubuk, Zoph, Shlens, and Le]{cubuk2020randaugment}
Ekin~D Cubuk, Barret Zoph, Jonathon Shlens, and Quoc~V Le.
\newblock Randaugment: Practical automated data augmentation with a reduced
  search space.
\newblock In \emph{Proceedings of the IEEE/CVF Conference on Computer Vision
  and Pattern Recognition Workshops}, pages 702--703, 2020.

\bibitem[Yun et~al.(2019)Yun, Han, Oh, Chun, Choe, and Yoo]{yun2019cutmix}
Sangdoo Yun, Dongyoon Han, Seong~Joon Oh, Sanghyuk Chun, Junsuk Choe, and
  Youngjoon Yoo.
\newblock Cutmix: Regularization strategy to train strong classifiers with
  localizable features.
\newblock In \emph{Proceedings of the IEEE/CVF international conference on
  computer vision}, pages 6023--6032, 2019.

\bibitem[Zhang et~al.(2017)Zhang, Cisse, Dauphin, and
  Lopez-Paz]{zhang2017mixup}
Hongyi Zhang, Moustapha Cisse, Yann~N Dauphin, and David Lopez-Paz.
\newblock mixup: Beyond empirical risk minimization.
\newblock \emph{arXiv preprint arXiv:1710.09412}, 2017.

\bibitem[Zhong et~al.(2020)Zhong, Zheng, Kang, Li, and Yang]{zhong2020random}
Zhun Zhong, Liang Zheng, Guoliang Kang, Shaozi Li, and Yi~Yang.
\newblock Random erasing data augmentation.
\newblock In \emph{Proceedings of the AAAI conference on artificial
  intelligence}, 2020.

\bibitem[Wightman et~al.(2021)Wightman, Touvron, and
  J{\'e}gou]{wightman2021resnet}
Ross Wightman, Hugo Touvron, and Herv{\'e} J{\'e}gou.
\newblock Resnet strikes back: An improved training procedure in timm.
\newblock \emph{arXiv preprint arXiv:2110.00476}, 2021.

\end{thebibliography}
}

\clearpage
\appendix
\section{Detailed architecture of \arch}
\label{sec:app_architecture}

\begin{figure}[b!]
    \centering
    \resizebox{\columnwidth}{!}{
        \hspace{-50px}\input{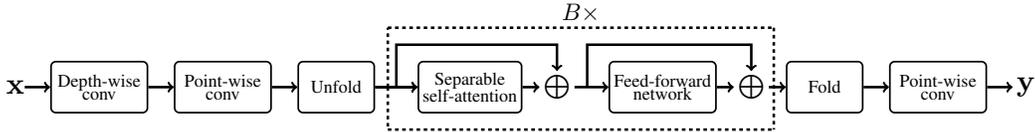}\mvtBlock
    }
    \caption{\textbf{\arch block}. Here, depth-wise convolution uses a kernel size of $3\times 3$ to encode local representations. Similar to \cite{mehta2022mobilevit}, unfolding and folding operations uses a patch height and width of two respectively. The separable self-attention and feed-forward layers are repeated $B \times$ before applying the folding operation.}
    \label{fig:arch_block}
\end{figure}

\arch's architecture follows MobileViTv1 \cite{mehta2022mobilevit} and is given in Table \ref{tab:app_arch}. \arch block, shown in \cref{fig:arch_block}, makes two changes to the MobileViTv1 block: (1) it replaces the multi-headed self-attention with the proposed separable self-attention to learn global representations and (2) it does not use fusion block and skip-connection (see Fig. 1b in \cite{mehta2022mobilevit}) as they improve the performance marginally (see Fig. 12 in \cite{mehta2022mobilevit}). The expansion factor in MobileNetv2 \cite{sandler2018mobilenetv2} blocks and feed-forward layers is two. Similar to \cite{mehta2022mobilevit}, we use Swish \cite{elfwing2018sigmoid} as a non-linear activation function. Unlike MobileViTv1 that creates three specific architectures (XXS, XS, and S) for mobile devices, we uniformly scale the width of \arch~network using a width multiplier $\alpha \in {0.5, 2.0}$ to create models at different complexities.

\begin{table}[b!]
    \centering
    \caption{\textbf{\arch~architecture.} Here, $d$ represents dimensionality of the input to the separable self-attention layer, $B$ denotes the repetition of transformer block with separable self-attention inside the \arch~block (\cref{fig:arch_block}), and MV2 indicates MobileNetv2 block. Similar to MobileViTv1 block, we set kernel size as three and spatial dimensions of patch (height $h$ and width $w$) as two in the \arch~block.}
    \label{tab:app_arch}
    \resizebox{0.85\columnwidth}{!}{
        \begin{tabular}{lcccl}
            \toprule[1.5pt]
            \textbf{Layer} & \textbf{Output size} & \textbf{Output stride} & \textbf{Repeat} & \textbf{Output channels} \\
            \midrule[1pt]
            Image & $256 \times 256$ & 1 & \multicolumn{2}{c}{} \\
            \midrule
            Conv-$3\times 3$, $\downarrow 2$ & \multirow{2}{*}{$128 \times 128$} & \multirow{2}{*}{2} & 1 & $32 \alpha$\\
            MV2 & & & 1 & $64 \alpha$  \\
            \midrule
            MV2, $\downarrow 2$ & \multirow{2}{*}{$64 \times 64$} & \multirow{2}{*}{4} & 1 & $128 \alpha$ \\
            MV2 & &  & 2 & $128 \alpha$ \\
            \midrule
            MV2, $\downarrow 2$ & \multirow{2}{*}{$32 \times 32$} & \multirow{2}{*}{8} & 1 & $256 \alpha$ \\
            \arch~block (\cref{fig:arch_block}; $B=2$) & &  & 1 & $256 * \alpha$ ($d = 128 \alpha)$ \\
            \midrule
            MV2, $\downarrow 2$ & \multirow{2}{*}{$16 \times 16$} & \multirow{2}{*}{16} & 1 & $384 \alpha$ \\
            \arch~block (\cref{fig:arch_block}; $B=4$) & &  & 1 &  $384 \alpha$ ($d = 192 \alpha)$ \\
            \midrule
            MV2, $\downarrow 2$ & \multirow{3}{*}{$8 \times 8$} & \multirow{3}{*}{32} & 1 & $512 \alpha$  \\
            \arch~block (\cref{fig:arch_block}; $B=3$) & &  & 1 & $512 \alpha$ ($d = 256 \alpha)$ \\
            \midrule
            Global pool & \multirow{2}{*}{$1 \times 1$} & \multirow{2}{*}{256} & \multirow{2}{*}{1} &  $512 \alpha$\\
            Linear &  &  &  & 1000 \\
            \bottomrule[1.5pt]
        \end{tabular}
    }
\end{table}

\section{\arch's classification performance} 
\label{sec:app_mvitv2_in1k}

\paragraph{ImageNet-1k} \cref{tab:mv2_pretrain_finetune} shows the results of \arch on the ImageNet-1k dataset. Finetuning \arch models at higher resolution ($384 \times 384$) shows improvement across the board. For example, the performance of \arch-0.50 with 1.4 million parameters improves by about 2\% when finetuned at higher resolution (R1 vs. R2). Similarly, pre-training on the ImageNet-21k-P dataset helps improve the performance of \arch~models. For example, ImageNet-21k-P pretraining improves the performance of \arch-2.0 improves by 1.2\% (R17 vs. R18). Notably, \arch models pretrained on the ImageNet-21k-P are able to achieve the similar performance with fewer FLOPs to models finetuned on ImageNet-1k with a higher resolution (e.g., R10 vs. R11; R14 vs. R15; R18 vs. R19 in \cref{tab:mv2_pretrain_finetune}).

\paragraph{ImageNet-21k-P} Table \ref{tab:app_imagenet21k_p_res} shows the results on the ImageNet-21k-P validation dataset. The performance of \arch improves with increase in model size. 

\begin{table}[t!]
    \centering
    \caption{\textbf{Classification performance of \arch on the ImageNet-1k dataset.} Here, $^\dagger$ indicates finetuning at higher resolution.}
    \label{tab:mv2_pretrain_finetune}
    \resizebox{0.75\columnwidth}{!}{
        \begin{tabular}{lcccrrc}
            \toprule[1.5pt]
           \textbf{Row \#} & \textbf{Model} & \textbf{Image size} & \textbf{Extra data} & \textbf{\# Params \contour{black}{$\downarrow$} } & \textbf{FLOPs \contour{black}{$\downarrow$}} & \textbf{Top-1 \contour{black}{$\uparrow$} } \\
            \midrule[1.25pt]
            R1 & \arch-0.50 & $256^2$ & None &  1.4 M & 0.5 G & 70.2 \\
            R2 & \arch-0.50$^\dagger$ & $384^2$ & None &  1.4 M & 1.0 G & 72.1 \\
            \midrule
            R3 & \arch-0.75 & $256^2$ & None &  2.9 M & 1.0 G & 75.6 \\
            R4 & \arch-0.75$^\dagger$ & $384^2$ & None & 2.9 M & 2.3 G & 77.0 \\
            \midrule
            R5 & \arch-1.00 & $256^2$ & None &  4.9 M & 1.8 G & 78.1 \\
            R6 & \arch-1.00$^\dagger$ & $384^2$ & None &  4.9 M & 4.1 G & 79.7 \\
            \midrule
            R7 & \arch-1.25 & $256^2$ & None &  7.5 M & 2.8 G & 79.6 \\
            R8 & \arch-1.25$^\dagger$ & $384^2$ & None &  7.5 M & 6.3 G & 80.9 \\
            \midrule
            R9 & \arch-1.50 & $256^2$ & None &  10.6 M & 4.0 G & 80.4 \\
            R10 & \arch-1.50 & $256^2$ & ImageNet-21k-P &  10.6 M & 4.0 G & 81.5 \\
            R11 & \arch-1.50$^\dagger$ & $384^2$ & None &  10.6 M & 9.1 G & 81.5 \\
            R12 & \arch-1.50$^\dagger$ & $384^2$ & ImageNet-21k-P &  10.6 M & 9.1 G & 82.6 \\
            \midrule
            R13 & \arch-1.75 & $256^2$ & None &  14.3 M & 5.5 G & 80.8 \\
            R14 & \arch-1.75 & $256^2$ & ImageNet-21k-P &  14.3 M & 5.5 G & 81.9 \\
            R15 & \arch-1.75$^\dagger$ & $384^2$ & None &  14.3 M & 12.3 G  & 82.0 \\
            R16 & \arch-1.75$^\dagger$ & $384^2$ & ImageNet-21k-P &  14.3 M & 12.3 G & 82.9 \\
            \midrule
            R17 & \arch-2.00 & $256^2$ & None &  18.5 M & 7.2 G & 81.2 \\
            R18 & \arch-1.75 & $256^2$ & ImageNet-21k-P &  18.5 M & 7.2 G & 82.4 \\
            R19 & \arch-2.00$^\dagger$ & $384^2$ & None &  18.5 M & 16.1 G & 82.2 \\
            R20 & \arch-1.50$^\dagger$ & $384^2$ & ImageNet-21k-P &  18.5 M & 16.1 G & 83.4  \\
            \bottomrule[1.5pt]
        \end{tabular}
    }
\end{table}

\begin{table}[t!]
    \centering
    \caption{Performance of \arch on the ImageNet-21k-P validation set.}
    \label{tab:app_imagenet21k_p_res}
    \resizebox{0.4\columnwidth}{!}{
        \begin{tabular}{ccccc}
            \toprule[1.5pt]
            \textbf{Width factor $\alpha$} & \textbf{\# Params \contour{black}{$\downarrow$}} & \textbf{FLOPs \contour{black}{$\downarrow$}} & \textbf{Top-1 \contour{black}{$\uparrow$}} & \textbf{Top-5 \contour{black}{$\uparrow$}} \\
             \midrule[1.25pt]
             1.50 & 17.9 M & 4.1 G & 44.5 & 74.5\\
             1.75 & 22.7 M & 5.5 G & 45.8 & 75.8 \\
             2.00 & 28.1 M & 7.2 G & 46.4 & 76.6\\
             \bottomrule[1.5pt]
        \end{tabular}
    }
\end{table}

\section{Comparisons with light-weight networks on the ImageNet-1k dataset}
\label{sec:app_mvitv2_light}

\begin{figure}[t!]
    \centering
    \begin{subfigure}[b]{0.42\columnwidth}
        \centering
        \includegraphics[width=0.8\columnwidth]{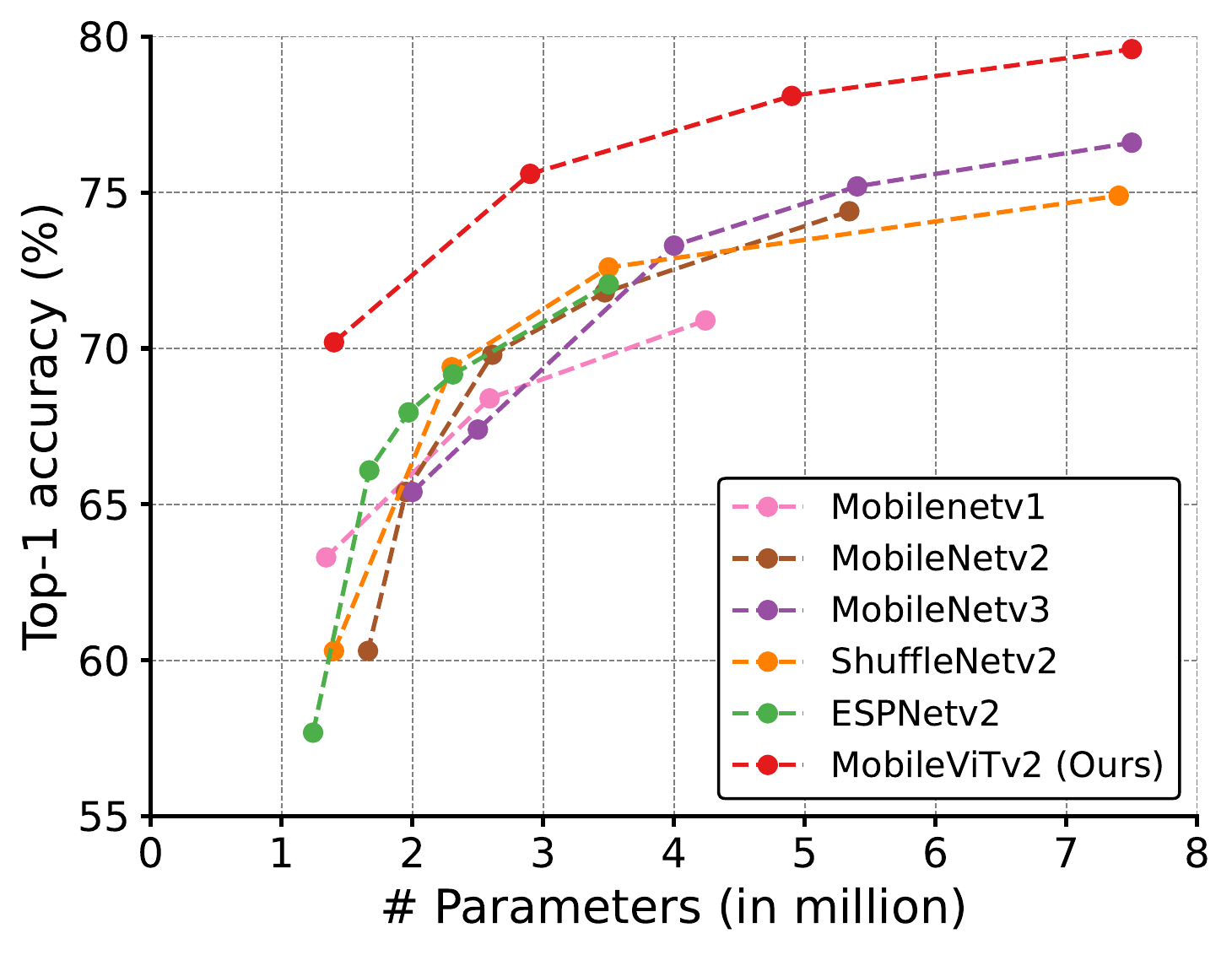}
        \caption{Comparison with light-weight CNNs}
        \label{fig:compare_imagenet_cnn_graph}
    \end{subfigure}
    \hfill
    \begin{subfigure}[b]{0.42\columnwidth}
        \centering
        \includegraphics[width=0.8\columnwidth]{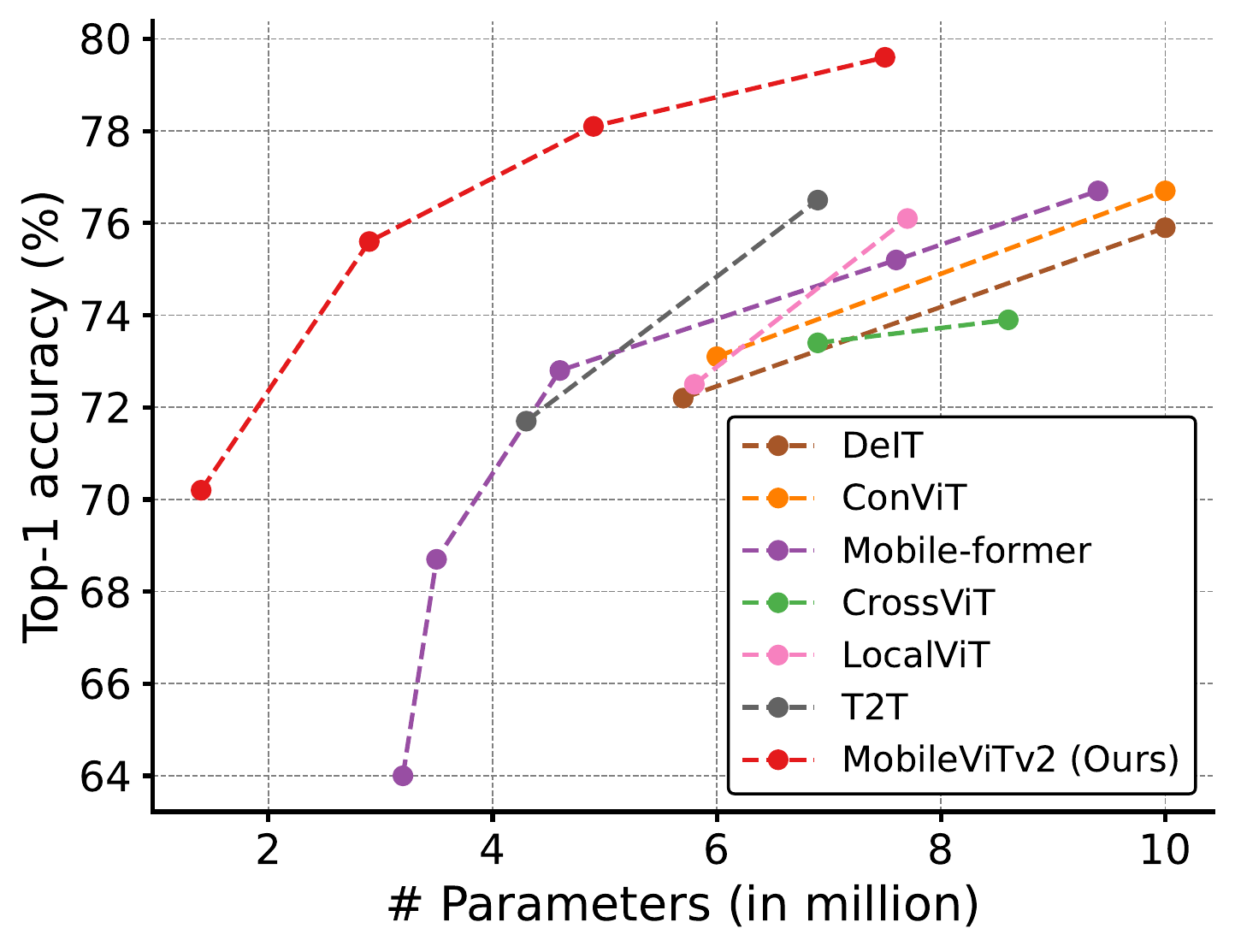}
        \caption{Comparison with light-weight ViTs}
        \label{fig:compare_imagenet_transformer_graph}
    \end{subfigure}
    \caption{\textbf{Comparison with light-weight CNN- and ViT-based models.} \arch~is smaller and better, which is desirable for mobile devices.}
    \label{fig:comapre_light_cnn_vit}
\end{figure}

\textbf{Comparison with light-weight CNNs.}  \cref{fig:compare_imagenet_cnn_graph} shows that \arch~outperforms light-weight CNNs across different network sizes (MobileNetv1 \citep{howard2017mobilenets}, MobileNetv2 \citep{sandler2018mobilenetv2}, ShuffleNetv2 \citep{ma2018shufflenet}, ESPNetv2 \citep{mehta2019espnetv2}, and MobileNetv3 \citep{howard2019searching}). 

\paragraph{Comparison with light-weight ViTs.}  \cref{fig:compare_imagenet_transformer_graph} shows that \arch~achieves better performance than previous light-weight ViT-based models acorss different network sizes (DeIT \citep{touvron2021training}, T2T \citep{yuan2021tokens}, CrossViT \citep{chen2021crossvit}, LocalViT \citep{li2021localvit}, ConViT \citep{d2021convit},  and Mobile-former \citep{chen2021mobile}).

\begin{figure}[t!]
    \centering
    \begin{subfigure}[b]{0.495\columnwidth}
        \resizebox{\columnwidth}{!}{
            \begin{tabular}{c}
                \toprule[1.5pt]
               \includegraphics[height=120px]{attn_map_vis/sample_1/000000012639.jpg} \\
               Input Image \\
               \midrule
               \includegraphics[height=65px]{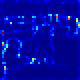}   \includegraphics[height=65px]{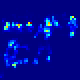} \\
               Context score map $\mathbf{c_m}$ at an output stride of 8 \\
               \midrule
               \includegraphics[height=65px]{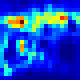}  \includegraphics[height=65px]{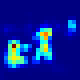} \includegraphics[height=65px]{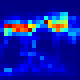}  \includegraphics[height=65px]{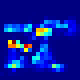} \\
               Context score map $\mathbf{c_m}$ at an output stride of 16 \\
               \midrule
               \includegraphics[height=65px]{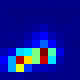}  \includegraphics[height=65px]{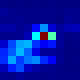}  \includegraphics[height=65px]{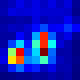} \\
               Context score map $\mathbf{c_m}$ at an output stride of 32 \\
               \bottomrule[1.5pt]
            \end{tabular}
        }
        \caption{}
    \end{subfigure}
    \hfill
    \begin{subfigure}[b]{0.495\columnwidth}
        \resizebox{\columnwidth}{!}{
            \begin{tabular}{c}
                \toprule[1.5pt]
               \includegraphics[height=120px]{attn_map_vis/sample_2/000000017029.jpg} \\
               Input Image \\
               \midrule
               \includegraphics[height=65px]{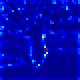}   \includegraphics[height=65px]{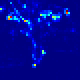} \\
               Context score map $\mathbf{c_m}$ at an output stride of 8 \\
               \midrule
               \includegraphics[height=65px]{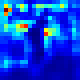}  \includegraphics[height=65px]{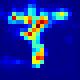} \includegraphics[height=65px]{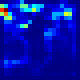}  \includegraphics[height=65px]{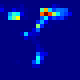} \\
               Context score map $\mathbf{c_m}$ at an output stride of 16 \\
               \midrule
               \includegraphics[height=65px]{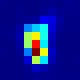}  \includegraphics[height=65px]{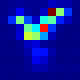}  \includegraphics[height=65px]{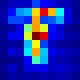} \\
               Context score map $\mathbf{c_m}$ at an output stride of 32 \\
               \bottomrule[1.5pt]
            \end{tabular}
        }
        \caption{}
    \end{subfigure}
    \vfill
     \begin{subfigure}[b]{0.495\columnwidth}
        \resizebox{\columnwidth}{!}{
            \begin{tabular}{c}
                \toprule[1.5pt]
               \includegraphics[height=120px]{attn_map_vis/sample_3/000000074733.jpg} \\
               Input Image \\
               \midrule
               \includegraphics[height=65px]{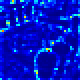}   \includegraphics[height=65px]{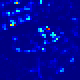} \\
               Context score map $\mathbf{c_m}$ at an output stride of 8 \\
               \midrule
               \includegraphics[height=65px]{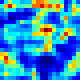}  \includegraphics[height=65px]{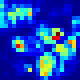} \includegraphics[height=65px]{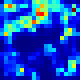}  \includegraphics[height=65px]{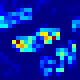} \\
               Context score map $\mathbf{c_m}$ at an output stride of 16 \\
               \midrule
               \includegraphics[height=65px]{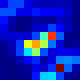}  \includegraphics[height=65px]{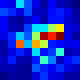}  \includegraphics[height=65px]{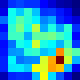} \\
               Context score map $\mathbf{c_m}$ at an output stride of 32 \\
               \bottomrule[1.5pt]
            \end{tabular}
        }
        \caption{}
     \end{subfigure}
     \hfill
     \begin{subfigure}[b]{0.495\columnwidth}
        \resizebox{\columnwidth}{!}{
            \begin{tabular}{c}
                \toprule[1.5pt]
               \includegraphics[height=120px]{attn_map_vis/sample_4/000000094871.jpg} \\
               Input Image \\
               \midrule
               \includegraphics[height=65px]{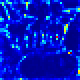}   \includegraphics[height=65px]{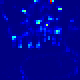} \\
               Context score map $\mathbf{c_m}$ at an output stride of 8 \\
               \midrule
               \includegraphics[height=65px]{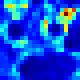}  \includegraphics[height=65px]{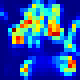} \includegraphics[height=65px]{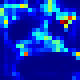}  \includegraphics[height=65px]{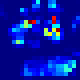} \\
               Context score map $\mathbf{c_m}$ at an output stride of 16 \\
               \midrule
               \includegraphics[height=65px]{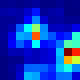}  \includegraphics[height=65px]{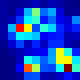}  \includegraphics[height=65px]{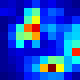} \\
               Context score map $\mathbf{c_m}$ at an output stride of 32 \\
               \bottomrule[1.5pt]
            \end{tabular}
        }
        \caption{}
     \end{subfigure}
    \caption{Layer-wise visualization of context score maps $\mathbf{c_m}$ at different output strides. Recall that \arch (\cref{fig:arch_block} and \cref{tab:app_arch}) applies $B=2$, $B=4$, and $B=3$ separable self-attention layers at an output strides of 8, 16, and 32 respectively. Therefore, we have 2, 4, and 8 context score maps at an output stride of 8, 16, and 32 respectively }
    \label{fig:context_score_maps}
\end{figure}

\section{Visualizations of separable self-attention scores}
\label{sec:app_vis}
The \arch block, \cref{fig:arch_block}, unfolds the input $\mathbf{x} \in \mathbb{R}^{d \times H \times W}$ to obtain $\mathbf{x_u} \in \mathbb{R}^{d, \times M \times N}$, where $N=\frac{HW}{hw}$ are the number of patches, each patch with width $w$ and height $h$ ($M=hw$ pixels per patch). This unfolded feature map is fed to separable self-attention module to learn non-local representations. To better understand how separable self-attention processes $\mathbf{x_u}$, we visualize context scores $\mathbf{c_s}$. 

The separable self-attention in \arch block computes context scores $\mathbf{c_s}$ for $M$ pixels simultaneously across $N$ patches. Therefore, $\mathbf{c_s}$ has a dimensions of $M \times N$. To visualize context scores, we fold  $\mathbf{c_s} \in \mathbb{R}^{M \times N}$ to the same spatial dimensions as the input and obtain context score map $\mathbf{c_m} \in \mathbb{R}^{H \times W}$. For ease of visualization, we scale $\mathbf{c_m}$ using min-max normalization. 

The context score maps for different input images at different output strides of \arch model are shown in \cref{fig:context_score_maps}. These visualizations show that the proposed separable self-attention method is able to (1) aggregate information from entire image under different settings, including complex backgrounds, illumination \& view-point changes, and different objects, and (2) learn high-, mid-, and low-level representations.

\section{\arch's ablation studies on the ImageNet-1k dataset}
\label{sec:app_mvitv2_ablate}
In this section, we study the effect on different methods on the performance of \arch~models, including augmentation methods.

\begin{figure}[b!]
    \centering
    \includegraphics[width=0.4\columnwidth]{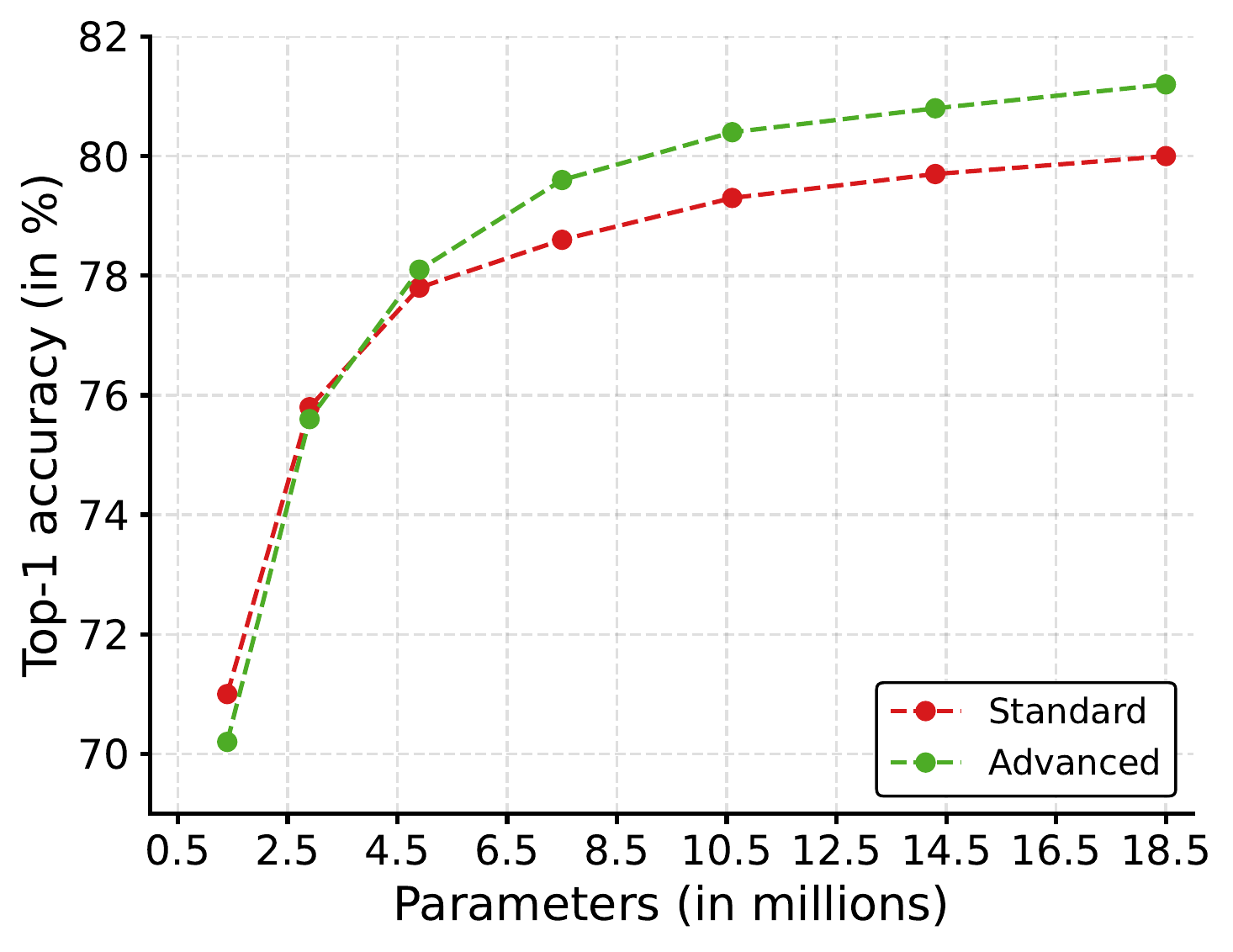}
    \caption{\textbf{Impact of data augmentation on the performance of \arch models on the ImageNet-1k dataset.} For smaller  models ($<4.5$ million parameters), standard augmentation works best while larger models ($\ge 4.5$ million parameters) benefit from advanced augmentation.}
    \label{fig:aug_res}
\end{figure}

\paragraph{Standard vs. advanced augmentation} We study two different augmentation methods: (1) standard augmentation that uses Inception-style augmentation \cite{szegedy2015going}, i.e., random resized cropping and horizontal flipping and (2) advanced augmentation that uses RandAugment \cite{cubuk2020randaugment}, CutMix \cite{yun2019cutmix}, MixUp \cite{zhang2017mixup}, and RandomErase \cite{zhong2020random} along with standard augmentation methods. The effect of these augmentations on the performance of \arch is shown in Figure \ref{fig:aug_res}. Smaller models ($<4.5$ million parameters) benefit from standard augmentation while larger models ($\ge 4.5$ million parameters) benefit from advanced augmentation. For simplicity, we use advanced augmentation for all variants of \arch in this paper. 

\paragraph{Loss functions} CutMix and Mixup augmentations mixes the samples in a batch. As a result, each sample has multiple labels. Therefore, in presence of these augmentations, ImageNet classification can be thought as a multi-label classification task. Similar to \cite{wightman2021resnet}, we trained \arch by minimizing binary cross-entropy loss. Unlike \cite{wightman2021resnet}, we did not observe any improvements in the performance when cross-entropy loss with label smoothing is replaced with binary cross-entropy loss. Therefore, we use cross-entropy with label smoothing for training \arch models.  

\paragraph{Effect of multiple latent tokens} Similar to multi-head attention in transformers, the proposed separable self-attention can have multiple latent tokens. When we changed the number of latent tokens from $1$ to $8$, the performance improvements on the ImageNet-1k dataset were negligible (within $\pm 0.1$ top-1 accuracy). Therefore, we use only one latent token in our experiments.

We note that changing the number of heads from $4$ to $1$ in multi-headed self-attention in the transformer block of the MobileViTv1-S architecture dropped the top-1 accuracy by 0.7\%. This observation is similar to \citet{vaswani2017attention}, who also found that multiple heads in multi-headed self-attention improve transformers performance on the task of neural machine translation.

\paragraph{Improving FLOP-efficiency via pixel- and patch-sampling} The MobileViTv1 model \cite{mehta2022mobilevit} unfolds an input feature map into $N$ patches, each patch with $M=hw$ pixels and applies a transformer block for each pixel in a patch independently, where $h$ and $w$ are patch's height and width respectively. Because pixels in a patch are spatially correlated, one can sub-sample $m$ pixels from $M$ pixels and learn non-local representations by applying self-attention layers on $m$ pixels only. Such sub-sampling methods should help in reducing model FLOPs. 

We tried following sampling methods at pixel- as well as patch-level: 
\begin{itemize}[leftmargin=*]
    \item \textbf{Random sampling}, wherein $m$ pixels (or $n$ patches) from $M$ pixels (or $N$ patches) are randomly selected during training and uniformly during validation.
    \item \textbf{Top-$m$ (or top-$n$) sampling}, wherein top-$m$ pixels (or top-$n$ patches) are selected based on their magnitude computed using L2 norm. 
    \item \textbf{Uniform sampling}, wherein $m$ pixels (or $n$ patches) are sampled uniformly from $M$ pixels (or $N$ patches). 
\end{itemize}    

We found that these methods can reduce the FLOPs by $1.2\times$ to $1.6 \times$ with little or no drop in top-1 accuracy on the ImageNet-1k dataset for both MobileViTv1 (with multi-headed self-attention) and \arch (with the proposed separable self-attention) models. However, these improvements in FLOPs did not translate to latency improvements on a mobile device. In fact, models with these sampling methods were significantly slower than the models without these methods. The high-latency of models with these sampling methods on mobile devices can be attributed to their high memory access cost, as these methods change the memory order of tensor. Because of their high-latency on mobile devices, we did not use these methods in the \arch model.

\section{\arch training configurations}

Configurations for training and finetuning \arch-2.0 on the ImageNet-1k and ImageNet-21k-P datasets are given in \cref{tab:app_pretrain_recipe} and \cref{tab:app_finetune_recipe} respectively while configurations for finetuning \arch on downstream tasks are given in \cref{tab:app_receipe_downstream_task}.

\begin{table}[b!]
    \centering
    \resizebox{0.55\columnwidth}{!}{
        \begin{tabular}{lcc}
        \toprule[1.5pt]
           \textbf{Training config}  & \multicolumn{2}{c}{\arch-2.0} \\
           \midrule[1.25pt]
           Dataset & ImageNet-1k & ImageNet-21k-P \\
           \# Training samples & 1.28 M & 11 M \\
           \# Validation samples & 50 k & 523 k$^\dagger$ \\
           Train resolution & $256 \times 256$ & $256 \times 256$ \\
           Val resolution &  $256 \times 256$ & $256 \times 256$ \\
           \midrule
           RandAug & \cmark & \cmark \\
           CutMix & \cmark & \cmark \\
           MixUp & \cmark & \cmark \\
           Random resized crop & \cmark & \cmark \\
           Random horizontal flip & \cmark & \cmark\\ 
           Random erase & \cmark & \cmark \\
           Stochastic depth & \xmark & \xmark \\
           \midrule
            Label smoothing & \cmark & \cmark \\
            Loss & CE & CE \\
            Optimizer & AdamW & AdamW \\
            Weight decay & 0.05 & 0.05 \\
            \midrule
            Scheduler & Cosine & Cosine \\
            Warm-up iterations & 20 k & None \\
            Warm-up init LR & $1e^{-6}$ & None \\
            Warm-up scheduler & Linear & None \\
            Base LR & 0.002 & 0.0003 \\
            Epochs & 300 & 80 \\
            Batch size & 1024 & 4096 \\
            Layer-wise LR decay & \xmark & \xmark \\
            \midrule
            Grad. clip & 10 & 10 \\
            Exp. moving average & \cmark & \cmark \\
            Weight init & Random & ImageNet-1k \\
            \bottomrule[1.5pt]
        \end{tabular}
    }
    \caption{Configuration for training \arch-2.0 on the ImageNet-1k/22k-P datasets. $^\dagger$ The validation set in ImageNet-21k-P does not overlap with ImageNet-1k validation set, and is created following \citet{ridnik2021imagenet}.}
    \label{tab:app_pretrain_recipe}
\end{table}

\begin{table}[b!]
    \centering
    \resizebox{0.7\columnwidth}{!}{
        \begin{tabular}{lccc}
        \toprule[1.5pt]
           \textbf{Training config}  & \multicolumn{3}{c}{\arch-2.0} \\
           \midrule[1.25pt]
           Dataset & ImageNet-1k & ImageNet-1k & ImageNet-1k\\
           \# Training samples & 1.28 M & 1.28 M & 1.28 M \\
           \# Validation samples & 50 k & 50 k & 50 k \\
           Train resolution & $384 \times 384$ & $256 \times 256$ & $384 \times 384$\\
           Val resolution &  $384 \times 384$ & $256 \times 256$ & $384 \times 384$\\
           Weight init & ImageNet-1k & ImageNet-21k-P & ImageNet-21k-P-1k$^\dagger$ \\
           \midrule
           RandAug & \xmark & \cmark & \cmark\\
           CutMix & \xmark & \cmark & \cmark\\
           MixUp & \xmark & \cmark & \cmark\\
           Random resized crop & \cmark & \cmark & \cmark \\
           Random horizontal flip & \cmark & \cmark & \cmark\\ 
           Random erase & \xmark & \cmark & \cmark \\
           Stochastic depth & \xmark & \xmark & \xmark \\
           \midrule
            Label smoothing & \cmark & \cmark & \cmark\\
            Loss & CE & CE & CE\\
            Optimizer & SGD & SGD & SGD \\
            Weight decay & $4e^{-5}$ & $4e^{-5}$ & $4e^{-5}$\\
            \midrule
            Scheduler & Fixed & Cosine & Fixed\\
            Warm-up iterations & None & None & None\\
            Warm-up init LR & None & None & None\\
            Warm-up scheduler & None & None & None\\
            Base LR & 0.001 & 0.01 & 0.001 \\
            Epochs & 10 & 50 & 10 \\
            Batch size & 128 & 256 & 128 \\
            Layer-wise LR decay & \xmark & \xmark & \xmark\\
            \midrule
            Grad. clip & 10 & 10 & 10 \\
            Exp. moving average & \cmark & \cmark & \cmark \\
            \bottomrule[1.5pt]
        \end{tabular}
    }
    \caption{Configuration for finetuning \arch-2.0 on the ImageNet-1k dataset. Here, $^\dagger$ denotes that the ImageNet-21k-P model finetuned on the ImageNet-1k dataset at $256\times 256$ image resolution is used for initializing the weights.}
    \label{tab:app_finetune_recipe}
\end{table}

\begin{table}[t!]
    \centering
    \resizebox{\columnwidth}{!}{
        \begin{tabular}{lccc}
        \toprule[1.5pt]
           \textbf{Training config}  & SSDLite-\arch-1.75 & DeepLabv3-\arch-1.75 & DeepLabv3-\arch-1.75 \\
           \midrule[1.25pt]
           Dataset & MS-COCO & ADE20k & PASCAL VOC 2012\\
           Extra Data & None & None & COCO \\
           Task & Detection & Segmentation & Segmentation \\
           \# Training samples & 117 k & 20 k & 128 k \\
           \# Validation samples & 5 k & 2 k & 1.45 k \\
           Train resolution & $320 \times 320$ & $512 \times 512$ & $512 \times 512$\\
           Val resolution &  $320 \times 320$ & Shortest side $512$ & Shortest side $512$\\
           Weight init & ImageNet-1k & ImageNet-1k & ImageNet-1k \\
           \midrule
           SSD Cropping & \cmark & \xmark & \xmark\\
           Photometric distortion & \cmark & \cmark & \cmark \\
           Random horizontal flip & \cmark & \cmark & \cmark\\
           Resize & \cmark & \xmark & \xmark\\
           Random short size resize & \xmark & \cmark & \cmark \\
           Random Crop & \xmark & \cmark & \cmark \\
           Random Gaussian blur & \xmark & \cmark & \cmark \\
           Random rotation & \xmark & \cmark & \cmark \\
           \midrule
            Loss & Smooth L1 + CE & CE & CE\\
            Optimizer & AdamW & SGD & AdamW \\
            Weight decay & 0.05 & $1e^{-4}$ & 0.05 \\
            \midrule
            Scheduler & Cosine & Cosine & Cosine\\
            Warm-up iterations & 500 & None & 500 \\
            Warm-up init LR & $9e^{-5}$ & None & $5e^{-5}$ \\
            Warm-up scheduler & Linear & None & Linear\\
            Base LR & 0.0009 & 0.02 & 0.0005 \\
            Epochs & 200 & 120 & 50 \\
            Batch size & 128 & 16 & 128 \\
            Layer-wise LR decay & \xmark & \xmark & \xmark\\
            \midrule
            Grad. clip & 10 & 10 & 10 \\
            Exp. moving average & \cmark & \cmark & \cmark \\
            \bottomrule[1.5pt]
        \end{tabular}
    }
    \caption{Configuration for finetuning \arch on downstream tasks. For Ade20k, we found that SGD was more stable as compared to AdamW across different \arch configurations, and therefore, we used SGD for finetuning on Ade20k dataset. The configurations for \arch with PSPNet are the same as Deeplabv3 on both PASCAL VOC and Ade20k datasets.}
    \label{tab:app_receipe_downstream_task}
\end{table}

\end{document}